\documentclass[sigconf]{acmart}

\AtBeginDocument{%
	\providecommand\BibTeX{{%
			\normalfont B\kern-0.5em{\scshape i\kern-0.25em b}\kern-0.8em\TeX}}}

\copyrightyear{2022}
\acmYear{2022}
\setcopyright{acmcopyright}
\acmConference[KDD '22]{Proceedings of the 28th ACM SIGKDD Conference on Knowledge Discovery and Data Mining}{August 14--18, 2022}{Washington, DC, USA}
\acmBooktitle{Proceedings of the 28th ACM SIGKDD Conference on Knowledge Discovery and Data Mining (KDD '22), August 14--18, 2022, Washington, DC, USA}
\acmPrice{15.00}
\acmDOI{10.1145/3534678.3539449}
\acmISBN{978-1-4503-9385-0/22/08}

\usepackage{multirow}
\usepackage{graphicx} 
\usepackage{stfloats}
\usepackage{subfigure} 
\usepackage{hyperref}
\usepackage{bm}
\usepackage{algorithm}
\usepackage{algorithmic}
\usepackage[framemethod=tikz]{mdframed}
\usepackage{color}
\usepackage[section]{placeins}
\newtheorem{definition}{Definition}

\begin{document}
	
	\title{Multi-View Clustering for Open Knowledge Base Canonicalization}
	
\newcommand\blfootnote[1]{%
	\begingroup
	\renewcommand\thefootnote{}\footnote{#1}%
	\addtocounter{footnote}{-1}%
	\endgroup
}

\author{Wei Shen}
\affiliation{
	\institution{TMCC, TKLNDST, College of Computer Science, Nankai University}
	\city{Tianjin}
	\country{China}
}
\email{shenwei@nankai.edu.cn}
\author{Yang Yang$^*$}
\affiliation{
	\institution{TMCC, TKLNDST, College of Computer Science, Nankai University}
	\city{Tianjin}
	\country{China}
}
\email{y2@mail.nankai.edu.cn}
\author{Yinan Liu}
\affiliation{
	\institution{TMCC, TKLNDST, College of Computer Science, Nankai University}
	\city{Tianjin}
	\country{China}
}
\email{liuyn@mail.nankai.edu.cn}
\renewcommand{\shortauthors}{Wei Shen, Yang Yang, \& Yinan Liu}

	\begin{abstract}
		Open information extraction (OIE) methods extract plenty of OIE triples <noun phrase, relation phrase, noun phrase> from unstructured text, 
		which compose large open knowledge bases (OKBs). 
		Noun phrases and relation phrases in such OKBs are not canonicalized, which leads to scattered and redundant facts. 
		It is found that two views of knowledge (i.e., a fact view based on the fact triple and a context view
		based on the fact triple’s source context) provide complementary information that is vital to the task of OKB canonicalization, 
		which clusters synonymous noun phrases and relation phrases into the same group and assigns them unique identifiers. 
		However, these two views of knowledge have so far been leveraged in isolation by existing works.
		In this paper, we propose CMVC, a novel unsupervised framework that leverages these two views of knowledge jointly for canonicalizing OKBs 
		without the need of manually annotated labels. 
		To achieve this goal, we propose a multi-view CH K-Means clustering algorithm to mutually reinforce the clustering of view-specific embeddings learned from each view by considering their different clustering qualities. 
		In order to further enhance the canonicalization performance,
		we propose a training data optimization strategy in terms of data quantity and data quality respectively
		in each particular view to refine the learned view-specific embeddings in an iterative manner.
		Additionally, we propose a Log-Jump algorithm to predict the optimal number of clusters in a data-driven way without requiring any labels.
		We demonstrate the superiority of our framework through extensive experiments on multiple real-world OKB data sets against state-of-the-art methods.
	\end{abstract}
	\maketitle
	\blfootnote{$^*$Corresponding author}
	\vspace{-7mm}
	\section{Introduction}
	Closed information extraction (CIE) \cite{chang2006survey} requires pre-specified ontology, 
	and has made a contribution to the development of curated knowledge bases (CKBs) 
	(e.g., YAGO \cite{yago} and Freebase \cite{bollacker2008freebase}). 
	Large CKBs contain millions of entities and hundreds of millions of relational facts about them, 
	which are usually stored in the form of triples. 
	Entities and relations in CKBs are generally canonicalized and well defined with unique identifiers. 
	CKBs play a fundamental role in many real knowledge-driven applications including 
	semantic search \cite{xiong2017explicit} and knowledge reasoning \cite{liu2021kompare}. 
	However, the construction of CKBs usually requires significant human involvement to pre-define the ontology, 
	which has limitations in its adaptability to new domains or new data.
	
	Open information extraction (OIE) has been proposed to solve this problem, such as Standford OIE \cite{angeli2015leveraging} and MinIE \cite{gashteovski2017minie}. 
	Without any pre-specified ontology,
	OIE methods extract a large number of OIE triples <noun phrase, relation phrase, noun phrase> from unstructured text 
	to compose large open knowledge bases (OKBs), which makes them highly adaptable.
	Such kind of notable OKBs include ReVerb \cite{fader2011identifying} and TextRunner \cite{Banko2007Open}.
	In general, the coverage and diversity of OKBs are much higher than CKBs.
	
	However, unlike CKBs, noun (relation) phrases in OKBs are not well canonicalized and lack unique identifiers for them.
	This may result in storage of scattered and redundant facts, which hinder their subsequent utilization in the downstream applications.
	Here, we show three sentences as an example.\vspace{0.2em}
	
	\textit{Donald J. Trump is Chairman of Trump University.}
	
	\quad \textit{Donald Trump is the CEO of Trump Organization.}
	
	\qquad \textit{Trump is currently the CEO of Trump Organization.} \vspace{0.2em}
	
	An OIE system can extract the following three OIE triples from these three sentences.\vspace{0.2em}
	
	\textit{<Donald J. Trump, is Chairman of, Trump University> }
	
	\quad \textit{<Donald Trump, is the CEO of, Trump Organization> }
	
	\qquad \textit{<Trump, is currently the CEO of, Trump Organization> }\vspace{0.2em}
	
	Unfortunately, it is unknown for machines that \textit{Donald J. Trump}, \textit{Donald Trump} and \textit{Trump} refer to the same entity. 
	This means that machines would not return all the available facts about this entity	
	when querying the term \textit{Donald Trump} over the above OIE triples.
	Additionally, it can be seen that the last two OIE triples are redundant facts, only one of which needs to be stored in practice. 
	
	To deal with this crucial issue, much attention has been paid to the task of OKB canonicalization, that is, 
	convert OIE triples in OKBs to their canonicalized form, which clusters synonymous noun (relation) phrases into a group. 
	Intuitively, OKB canonicalization helps to integrate knowledge and eliminate redundancy of OKBs, which could greatly benefit the downstream applications. 
	
	In spite of its importance, canonicalization of OKBs is still an unsolved problem.
	To tackle it, two categories of models have been explored, and they all regard it as a clustering problem and 
	cluster synonymous noun (relation) phrases into the same group. 
	Specifically, the first line of models \cite{galarraga2014canonicalizing, vashishth2018cesi, dash2021open, liu2021joint} mainly utilizes the fact view, i.e., knowledge embedded in the OIE fact triple, to cluster noun (relation) phrases with the same semantics.
	Although some of them \cite{vashishth2018cesi, dash2021open} need the context as their input, they use the context as the input of some third-party tools to generate side information utilized in the fact view, rather than leveraging the context as an independent view for canonicalizing OKBs.
	The second category of method \cite{lin2019canonicalization} primarily employs the context view, i.e., knowledge embedded in the source context where the OIE fact triple is extracted.
	In reality, these two views of knowledge characterize different aspects of noun (relation) phrases and are complementary to each other.
	Making use of them together would improve the model accuracy and robustness.
	However, these two views of knowledge have been leveraged in isolation so far.
	
	In this paper, we propose CMVC, a novel unsupervised framework for \underline{\textbf{C}}anonicalizing OKBs based on \underline{\textbf{M}}ulti-\underline{\textbf{V}}iew \underline{\textbf{C}}lustering
	by leveraging these two views of knowledge jointly. 
	The fact view characterizes the relational structural information of fact triples, in which entities are linked by relations.
	The context view captures the semantic distributional information of the source context where the fact triple occurs.
	These two views provide complementary information both of which are vital to OKB canonicalization.
	To combine knowledge from both views, we propose a multi-view CH K-Means clustering algorithm to mutually reinforce the clustering of view-specific embeddings learned 
	from each view by considering their different clustering qualities. 
	
	Intuitively, better embeddings learned from multiple views lead to better canonicalization result. 
	Consequently, the goal of each view is to extract high-quality embeddings based on the view-specific information. 
	Inspired by a prior work \cite{vashishth2018cesi}, we could collect some seed pairs of synonymous noun (relation) phrases automatically from external resources 
	without any human involvement and then leverage them as prior knowledge for high-quality encoding. 
	However, these automatically collected seed pairs are often limited and error-prone.
	To address this issue, we propose a training data optimization strategy in terms of data quantity and data quality respectively in each particular view 
	to further promote the learned view-specific embeddings iteratively. 
	In the fact view, we focus on optimizing the quantity of the training data and propose a data augmentation operator, 
	i.e., swap counterparts of the seed pairs in their involved fact triples, to derive additional augmented training data. 
	In the context view, we focus on optimizing the quality of the training data and propose an iterative clustering procedure, 
	which alternately performs clustering and embedding learning, to transfer prior knowledge from seed pairs to other pairs to construct more accurate training data. 
	These two training data optimization strategies in both views boost the quality of the learned view-specific embeddings significantly.
	
	Our major contributions can be summarized as follows:
	
	$\bullet$ This is the first unsupervised framework that combines knowledge from both the fact view and the context view for canonicalizing OKBs 
	via a novel multi-view CH K-Means clustering algorithm. 
	
	$\bullet$ In each view, we propose a training data optimization strategy in terms of data quantity and data quality respectively 
	to refine the learned view-specific embeddings in an iterative manner.
	
	$\bullet$ We propose a Log-Jump algorithm to predict the optimal number of clusters in a data-driven way without requiring any labels.
	
	$\bullet$ A thorough experimental study over three real-world OKB data sets shows that our framework outperforms all the baseline methods 
	for the task of OKB canonicalization. 
	
	\section{PRELIMINARIES AND NOTATIONS}
	In this section, we introduce some basic concepts and define the task of OKB canonicalization.
	
	In an OKB, an OIE triple is denoted by $t_i = <sub_i, rel_i, obj_i > $, 
	where $sub_i$ and $obj_i$ are noun phrases (NPs) and $rel_i$ is a relation phrase (RP). 
	The source text where the OIE triple $t_i$ is extracted, is denoted by $s_{t_i}$, which could be a sentence or a paragraph in the source article. 
	For an NP $sub_{i}$, we define its source context as $c_{sub_i}$, i.e., the source text $s_{t_i}$ with the NP $sub_i$ itself removed. 
	The source context of $rel_i$ ($obj_i$) could be defined in a similar manner.
	
	\vspace{-1mm}
	\begin{definition}
		[\textbf{OKB Canonicalization}] 
		Given a set of OIE triples $t_{i}$'s in an OKB and their corresponding source texts $s_{t_{i}}$'s, 
		the goal of this task is to cluster synonymous NPs referring to the same entity and synonymous RPs having the same semantic meaning into a group, 
		which converts these OIE triples to the canonicalized form.
	\end{definition}
	\vspace{-1mm}
	It can be seen from the task definition above that two views 
	(i.e., a fact view based on the OIE fact triples $t_{i}$'s and a context view based on their corresponding source texts $s_{t_{i}}$'s) 
	are provided as input and our target is to leverage these two views of knowledge jointly. 
	Based on the knowledge from a specific view $v$, we could learn the view-specific embedding $\bm{sub}_i^{(v)}$ for the NP $sub_i$, 
	which is introduced in details in Section \ref{fact_view} (i.e., the fact view) and Section \ref{context_view} (i.e., the context view), respectively. 
	For notation, we use the bold lowercase letter to represent the embedding in this paper. 
	The view-specific embedding $\bm{rel}_i^{(v)}$ ($\bm{obj}_i^{(v)}$) of ${rel}_i$ (${obj}_i $) could be derived in a similar manner. 
	There are totally two views in our framework, i.e., $v \in \{ 1, 2 \}$, where view $1$ represents the fact view and view $2$ represents the context view. 

	\section{THE FRAMEWORK CMVC\label{sec:cmvc}}
\begin{figure}[t] 
		\centering 
		\includegraphics[width=0.49\textwidth]{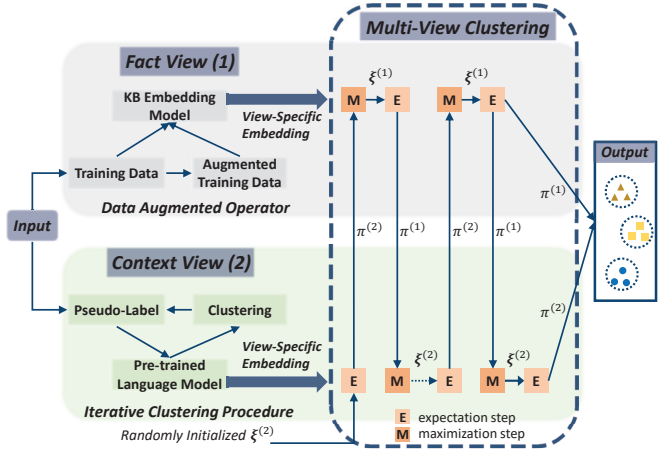} 
		\vspace{-5.5mm}
		\caption{An illustration of the proposed CMVC framework. }
		\vspace{-0.5mm}
		\label{figure_1} 
	\end{figure}

	The overall framework of our proposed CMVC is shown in Figure \ref{figure_1}. 
	We begin with the introduction of our proposed multi-view clustering algorithm and thereafter describe the fact view and the context view subsequently. 
	\vspace{-2mm}
	\subsection{Multi-View Clustering\label{sec:mvc}}
	\vspace{-1mm}
	Multi-view clustering has been widely studied in machine learning, 
	which aims to provide a more accurate and stable clustering result than single view clustering by considering complementary information of multiple views.  
	In this paper, we propose a novel multi-view CH K-Means algorithm for OKB canonicalization to integrate knowledge from two views (i.e., a fact view and a context view) 
	by considering their different clustering qualities. 
	In the following, first, we describe this proposed multi-view CH K-Means algorithm. 
	Next, we introduce our proposed novel Log-Jump algorithm, to predict the parameter (i.e., the number of clusters) of the clustering algorithm in a data-driven manner. 
	For simplicity, we take the NP $sub_{i}$ as an example for illustration. 
	The process for the RP $rel_{i}$ and NP $obj_{i}$ is similar and omitted for saving space. 
	
	\setlength{\textfloatsep}{0.7mm}
	\setlength{\floatsep}{0.7mm}
	
	\vspace{-1.5mm}
	\subsubsection{Multi-View CH K-Means\label{sec:mvch}}
	Inspired by the multi-view spherical K-Means \cite{bickel2004multi}, our multi-view CH K-Means is based on the co-EM algorithm. 
	In order to quantify the quality of the clustering result for each view, we leverage Cali$\rm{\acute{n}}$ski-Harabasz (CH) index \cite{calinski1974dendrite} as the weight of each view, instead of treating different views equally. 
	To mutually reinforce the clustering of each view, steps M, E in view 1 and steps M, E in view 2 are performed by turn 
	and the clustering result is transferred between two views. 
	We introduce the M-step and the E-step as follows.
	
	\textbf{\textit{M-step}}. The input of the maximization step is the clustering result $\pi^{(\bar{v})} = \{ \pi_{1}^{(\bar{v})}, ..., \pi_{j}^{(\bar{v})}, ..., \pi_{K}^{(\bar{v})} \}$ from the \textbf{\textit{other}} view $\bar{v}$ ($\bar{v}$ is a different view from view $v$), where $\pi_{j}^{(\bar{v})}$ is a cluster of NPs and $K$ is the desired number of clusters. 
	Then we calculate the cluster center embedding $\pmb{\xi}_{j}^{(v)}$ for the $j$-th cluster of view $v$ as follows: 
	\vspace{-2mm}
	\begin{equation}
	\vspace{-2mm}
	\pmb{\xi}_{j}^{(v)} = \frac { \sum\limits_{sub_{i} \in \pi_{j}^{(\bar{v})}} \bm{sub}_{i}^{(v)} }{ \| \sum\limits_{sub_{i} \in \pi_{j}^{(\bar{v})}} 
	\bm{sub}_{i}^{(v)}\| } \label{mvc_m}, 
	\vspace{-2mm}
	\end{equation}
	where $\bm{sub}_{i}^{(v)}$ is the view-specific embedding of view $v$ for the NP $sub_{i}$ and $\| \cdot \|$ denotes the $L_1$ vector norm. 
	
	\textbf{\textit{E-step}}. In the expectation step, given the cluster center embeddings $\pmb{\xi}^{(v)} = \{ \pmb{\xi}_{1}^{(v)}, ..., \pmb{\xi}_{j}^{(v)}, ..., \pmb{\xi}_{K}^{(v)} \}$ 
	of view $v$ calculated in the M-step, we assign an NP $sub_{i}$ to its corresponding cluster $\pi_{j}^{(v)}$ of view $v$ 
	by finding its closest cluster center embedding $ \pmb{\xi}_{j}^{(v)} $ based on the following formula: 
	\vspace{-2mm}
	\begin{equation}
	\begin{aligned}
	\pi_{j}^{(v)} = \{ sub_{i} \in S : sim( \bm{sub}_{i}^{(v)}, \pmb{\xi}_{j}^{(v)} ) \geq sim( \bm{sub}_{i}^{(v)}, \pmb{\xi}_{l}^{(v)} ) \}, \label{mvc_e}
	\end{aligned}
	\end{equation}
	where $l$, $j \in \{ 1, ... , K \} $, $l \neq j $, $sim( \cdot )$ is the cosine similarity, 
	and $S = \{ sub_1, ... , sub_i, ...  \} $ is the whole set of NPs. 
	It is noted that all the NP embeddings for the NP $sub_{i}$ in $S$ have unit length (i.e., $ \| \bm{sub}_{i}^{(v)} \| = 1 $). 
	
	After performing an M-step and an E-step in one view, 
	the clustering result will get interchanged for an M-step and an E-step in the other view, and so on, 
	which is shown in the right part of Figure \ref{figure_1}. 
	This iteration procedure is expected to minimize the loss function $\mathcal{L}_{mvc}$ as follows: 	
	\vspace{-1.5mm}
	\begin{equation}
	\mathcal{L}_{mvc} = \sum_{v = 1} ^{2} \sum_{j = 1} ^{K} \sum_{sub_{i} \in \pi_{j}^{(v)}} 1 - sim( \bm{sub}_{i}^{(v)}, \pmb{\xi}_{j}^{(v)} ), \label{mvc_loss} 
	\vspace{-1.5mm}
	\end{equation}
	When this iteration procedure converges, for each view $v$, we could yield a clustering result $\pi^{(v)}$. 
	However, there are still some conflicts between the clustering results of different views. 
	In order to eliminate conflicts and establish the final clustering result, 
	we assign each NP to one distinct cluster based on its closest cluster center embedding. 
	For this purpose, we calculate a consensus mean $\bm{m}_{j}^{(v)}$ for the $j$-th cluster of view $v$ as follows:
	\vspace{-2mm}
	\begin{equation}
	\vspace{-1mm}
	\bm{m}_{j}^{(v)} = \frac { \sum\limits_{sub_{i} \in \pi_{j}^{(1)} \land sub_{i} \in \pi_{j}^{(2)}} \bm{sub}_{i}^{(v)} }
	{ \| \sum\limits_{sub_{i} \in \pi_{j}^{(1)} \land sub_{i} \in \pi_{j}^{(2)}} \bm{sub}_{i}^{(v)}\| }, \label{mvc_mean}
	\vspace{-1mm}
	\end{equation}
	It can be seen that the calculation of the consensus mean $\bm{m}^{(v)}_j$ only considers those NPs that both views agree on. 
   Due to the fact that different views have different clustering qualities, we should treat them differently in the process of integrating their clustering results. 
   In order to quantify the quality of the clustering result for a view, we leverage a well-known statistics-based measure, i.e., Cali$\rm{\acute{n}}$ski-Harabasz (CH) index \cite{calinski1974dendrite}. 
   To be specific, we calculate the CH index of the clustering result $\pi^{(v)}$ of view $v$ as follows:
	\vspace{-1mm}
	\begin{equation}	
	\vspace{-1mm}
	{CH}^{(v)} = \frac { \sum_{j=1}^{K} \left| \pi_{j}^{(v)} \right| \cdot \| \pmb{\xi}_{j}^{(v)} - \pmb{\xi}^{(v)} \|^2 }
	{ \sum_{j=1}^{K} \sum\limits_{sub_{i} \in \pi_{j}^{(v)}} \| \bm{sub}_{i}^{(v)} - \pmb{\xi}_{j}^{(v)}\|^2 } \times \frac { \left| S \right| -K }{ K-1 }  \label{ch}, 
	\vspace{-1mm}
	\end{equation}
	where $\left| \pi_{j}^{(v)} \right|$ is the number of NPs in the cluster $\pi_{j}^{(v)}$, 
	$\pmb{\xi}^{(v)}$ is the global center embedding (i.e., the mean embedding) of all NP embeddings in view $v$, and $\left| S \right|$ is the total number of NPs. 
	From Formula \eqref{ch}, it can be seen that high value of the CH index means the clusters are well separated to each other and dense in each intra-cluster, 
	demonstrating that this CH index is a reasonable indicator for the clustering quality. 
	Therefore, we regard the CH index as the weight of each view and assign an NP $sub_{i}$ to the final cluster $ \hat{\pi_{j}}$ 
	by averaging over the weighted cosine similarities in both views based on the following formula:
	\vspace{-1mm}
	\begin{equation}
	\vspace{-1mm}
	\begin{aligned}
	\hat{\pi_{j}} = \{ sub_{i} \in S : \sum_{v = 1} ^{2} {CH}^{(v)} \cdot sim ( \bm{sub}_{i}^{(v)}, \bm{m}_{j}^{(v)} ) \\
	\geq \sum_{v = 1} ^{2} {CH}^{(v)} \cdot sim( \bm{sub}_{i}^{(v)}, \bm{m}_{l}^{(v)} ) \}, \label{mvc_pi}
	\end{aligned}
	\vspace{-1mm}
	\end{equation}
	where $l$, $j \in \{ 1, ... , K \} $ and $l \neq j $. 
	We define the final clustering result as $\hat{\pi} = \{ \hat{\pi_1}, ... ,  \hat{\pi_K} \} $. 
	This multi-view CH K-Means algorithm is summarized in Algorithm \ref{alg1} of Appendix \ref{sec:multi-view_ch}. 
	For the initialization, we randomly initialize the cluster center embeddings $\pmb{\xi}^{(2)}$ of view 2 as shown in Figure \ref{figure_1}. 
	
	\vspace{-1.5mm}
	\subsubsection{Data-Driven Parameter Prediction\label{sec:inverse-jump}}
	There is an important issue still unresolved, that is, the number of clusters (i.e., the parameter $K$) is unknown, which is an input of Algorithm \ref{alg1}. 
	For OKB canonicalization, data is large-scale and sparse, 
	and the number of clusters is usually large, which makes this issue challenging. 
	
	Previous OKB canonicalization studies \cite{galarraga2014canonicalizing, vashishth2018cesi, dash2021open} utilized the validation data set to 
	find the optimal clustering threshold in a semi-supervised manner. 
	Lin et al. \cite{lin2019canonicalization} set the clustering threshold of different data sets to the same fixed value in an unsupervised manner. 
	To remedy the issue in a more flexible and unsupervised manner, 
	we propose a novel algorithm called Log-Jump to predict the number of clusters in a data-driven way, 
	which only depends on the input embeddings of data without requiring any labels. 
	This Log-Jump algorithm is summarized in Algorithm \ref{alg2}, and we elaborate it as follows. 
	
	Given a set of input embeddings $X = \{ \pmb{x}_1, \pmb{x}_2, ... ,  \pmb{x}_n \} $ that needs to be grouped into clusters, $n$ is the number of input embeddings 
	and also the possible maximum number of clusters. 
	First, for each possible $K$, we derive the cluster center embeddings $\pmb{\xi} = \{ \pmb{\xi}_{1}, ..., \pmb{\xi}_{K} \}$ 
	via clustering the input embeddings $X$ using K-Means (w.r.t. lines 2 in Algorithm \ref{alg2}). 
	Subsequently, we calculate the distortion $d_{K}$ \cite{sugar2003findingtn} (w.r.t. lines 3 in Algorithm \ref{alg2}), 
	which is a quantity that measures the average distance, per dimension, 
	between each input embedding and its closest cluster center embedding and defined as follows:
	\vspace{-2mm}
	\begin{equation}
	\vspace{-1mm}
	d_{K} = \frac{1}{n \cdot p} \sum_{ \pmb{x}_{i} \in X } \min\limits_{j \in \{ 1,..,K \} } (1 - sim( \pmb{x}_{i}, \pmb{\xi}_{j} )), \label{inverse-jump-dk}
	\vspace{-2mm}
	\end{equation}
	where $p$ is the dimension of the input embedding, and $1 - sim( \pmb{x}_{i}, \pmb{\xi}_{j} )$ is the cosine distance between $\pmb{x}_{i}$ and $\pmb{\xi}_{j}$. 
	Next, for each possible $K$, we calculate its Log-Jump measure $LJ_{K}$ using a logarithmic function based on the calculated distortion 
	(w.r.t. line 6 in Algorithm \ref{alg2}). 
	In the end, we output the predicted number of clusters $K^{*}$, whose Log-Jump measure $LJ_{K}$ is the largest. 
	
	Overall, compared with the existing Jump method \cite{sugar2003findingtn}, our proposed Log-Jump algorithm not only 
	predicts the number of clusters based on the Log-Jump measure, 
	but also changes the manner of calculating the distortion $d_{K}$ from Mahalanobis distance to cosine distance. 
	We will give a theoretical justification for this method based on information theoretic ideas in Appendix \ref{A1} and 
	introduce a heuristic method to generate a small candidate range of possible $K$ for the sake of efficiency in Appendix \ref{A2}.
	\begin{algorithm}[t]
		\renewcommand{\algorithmicrequire}{\textbf{Input:}}
		\renewcommand{\algorithmicensure}{\textbf{Output:}}
		\caption{Log-Jump Algorithm}
		\label{alg2}
		\begin{algorithmic}[1]
			\REQUIRE A set of input embeddings $X = \{ \pmb{x}_1, ... ,  \pmb{x}_n \} $, the number of input embeddings $ n $. 
			\FOR{$K = 1$ to $ n $}{
				\STATE Compute the cluster center embeddings $\pmb{\xi} = K$-$Means(X, K)$\vspace{-3.5mm}
				\STATE Compute the distortion $ d_{K} $ by Formula \eqref{inverse-jump-dk}
			}
			\ENDFOR
			\FOR{$K = 1$ to $n - 1$}{
				\STATE $ LJ_{K} = \log d_{K+1} - \log d_{K} $ 
			}
			\ENDFOR
			\ENSURE  the predicted number of clusters $ K^{*} = \arg\max\limits_{K} {LJ_{K}} $
		\end{algorithmic}
	\end{algorithm}
	\vspace{-2mm}
	\subsection{Fact View\label{fact_view}}
	\vspace{-1mm}
	Based on the information from the fact view, we introduce how to learn view-specific embeddings of the fact view, called fact embeddings. 
	Next, we introduce how to promote the learned fact embeddings via the proposed training data optimization strategy (i.e., data augmentation operator).
	\vspace{-1.5mm}
	\subsubsection{Fact Embedding Learning}
	To capture the relational structural information of fact triples in an OKB, we could learn fact embeddings via a kind of KB embedding model (KEM), 
	such as TransE \cite{bordes2013translating} and RotatE \cite{ sun2018rotate}. 
	It is noted that any KEM could be employed in this framework, as long as it could encode fact triples into fact embeddings. 
	That is to say, given a fact triple $t_{i} = < sub_{i} , rel_{i} , obj_{i} >$ in an OKB, KEM can project $sub_{i}$, $rel_{i}$, and $obj_{i}$ 
	into fact embeddings $\bm{sub}_{i}^{(1)}$, $\bm{rel}_{i}^{(1)}$, and $\bm{obj}_{i}^{(1)}$ respectively. 
	Here, we assume the score function of the KEM is $f(\cdot)$, which measures the plausibility of a fact triple \cite{ji2021survey}. 
	To learn fact embeddings, KEM usually minimizes a margin-based loss function $\mathcal{L}_{fact}$ based on training data (i.e., fact triples): 
	\begin{equation}
	\vspace{-1mm}
	\mathcal{L}_{fact} = \sum_{t_i \in T^+} \sum_{t'_i \in T^-} \left[ \gamma + f(t_i) - f(t'_i) \right]_+, \label{transe_loss}
	\vspace{-1mm}
	\end{equation}
	where $ \left[ \alpha \right]_+ $ represents the maximum between $0$ and $\alpha$, 
	$ \gamma > 0 $ is a margin hyperparameter, 
	$ T^{+} $ denotes all the available fact triples of the OKB which are regarded as the positive training data, and
	$ T^{-} $ denotes the negative training data, which is usually generated via negative sampling and defined as follows:
	\begin{equation}
	\vspace{-0.5mm}
	\begin{aligned}
	T^{-} = \{ <sub'_i, rel_i, obj_i> | sub'_i \in N \} \cup \{ <sub_i, \\
	rel_i, obj'_i> | obj'_i \in N \}, 
	<sub_i, rel_i, obj_i> \in T^{+}, \label{negative_sample}
	\end{aligned}
	\vspace{-0.5mm}
	\end{equation}
	where $ N $ denotes the set of NPs in $ T^{+} $. 
	Formula \eqref{negative_sample} indicates that one of two NPs in a positive fact triple is replaced by another random NP (but not both at the same time). 
	
	\subsubsection{Data Augmentation Operator\label{sec:dao}}
	In general, fact triples of an OKB are usually sparse, which prevents the KEM from learning high-quality fact embeddings. 
	To make the training data dense and augment it with more instances, 
	we leverage prior knowledge embedded in seed pairs and propose a data augmentation operator, 
	i.e., swap counterparts of the seed pairs in their involved fact triples, 
	to derive additional augmented training data, 
	which could be regarded as a training data optimization strategy 
	in terms of data quantity. 
	For an OKB, the original training data (i.e., all the available fact triples in the OKB) is denoted by $ T^{O} $, 
	and the augmented training data is denoted by $ T^{A} $, 
	which is generated via the data augmentation operator. 
	As aforementioned, 
	we could collect some seed pairs of synonymous NPs automatically from external resources 
	without any human involvement 
	and thereafter leverage them as prior knowledge for high-quality encoding. 
	Let $(u_i, u_j)$ denote a seed pair, where $u_i$ and $u_j$ are synonymous NPs that refer to the same entity. 
	The set of collected seed pairs is denoted by $U$. Given the seed pair set $U= \{(u_{i}, u_{j}) \}$ 
	and the original training data $T^{O} = \{ <sub_{i},rel_{i},obj_{i}> \}$, 
	the augmented training data $ T^{A} $ is generated via swapping counterparts of the seed pairs in their involved fact triples as follows:  
	\vspace{-1.5mm}
	\begin{equation}
	\vspace{-1mm}
	\begin{aligned}
	T^{A} = \{ <u_j, rel_i, obj_i> | sub_i = u_i \} \cup \{ <sub_i, rel_i, u_j> | obj_i = u_i \} \\
	\cup \{ <u_i, rel_i, obj_i> | sub_i = u_j \} \cup \{ <sub_i, rel_i, u_i> | obj_i = u_j \}. \label{DAO}
	\end{aligned}
	\vspace{-1mm}
	\end{equation}
	\vspace{-1.5mm}
	
	Intuitively, a heuristic way to leverage the generated augmented training data $ T^{A} $ for encoding is combining $ T^{A} $ and $ T^{O} $ 
	into one positive training data set. 
	However, these automatically collected seed pairs are error-prone. 
	Therefore, triples in $ T^{A} $ are not entirely correct and the disagreement between $ T^{A} $ and $ T^{O} $ would not be helpful for learning high-quality embeddings. 
	To minimize the impact of the disagreement and further promote the learned fact embeddings, inspired by the idea of iterative training \cite{sun2018bootstrapping}, 
	we select $ T^{A} $ or $ T^{O} $ as the positive training data iteratively. 
	This way, KEM could learn superior fact embeddings based on the original training data $T^{O}$ and the augmented training data $ T^{A} $ alternately. 
	It should be noted that negative sampling also needs to be performed when $ T^{A} $ is selected as the positive training data. 
	\vspace{-2mm}
	\subsection{Context View\label{context_view}}
	\vspace{-1mm}
	Based on the information from the context view, we introduce how to extract view-specific embeddings of the context view, called context embeddings. 
	Next, we introduce how to refine the learned context embeddings 
	via the proposed training data optimization strategy 
	(i.e., iterative clustering procedure). 
	\vspace{-2mm}
	\subsubsection{Context Embedding Learning}
	To exploit the semantic distributional information of the source context where the fact triple occurs, 
	we extract context embeddings resorting to a kind of pre-trained language model (PLM), such as BERT \cite{kenton2019bert} and ELMo \cite{Peters:2018}. 
	For an NP $sub_i$, we leverage its source context $ c_{sub_i} $ as the input of the PLM and regard the output embedding as its context embedding $ \bm{sub}_{i}^{(2)} $. 
	It is also worth mentioning that any PLM could be leveraged in this framework, as long as it could encode source contexts into context embeddings.
	
	\vspace{-2.5mm}
	\subsubsection{Iterative Clustering Procedure\label{sec:icp}}
	When fine-tuned over enough task-specific labeled training data, PLM could achieve excellent performance on the specific task, 
	but these task-specific labeled training data usually require large amounts of manual annotation. 
	As source contexts in this task are unlabeled,
	we derive their pseudo-labels that represent the referring entities of their corresponding NPs as follows. 
	First, we cluster source contexts based on the seed pair set $U$ by grouping two source contexts into a cluster, 
	if their corresponding NPs form a pair in $U$, which means these two NPs are synonymous and refer to the same entity. 
	If an NP does not occur in any seed pair of $U$, its source context will be assigned into a singleton cluster. 
	In this case, source contexts grouped into the same group should have the same pseudo-label. 
	Based on this clustering result of source contexts, a distinct pseudo-label is assigned to each cluster, 
	and the source context in a cluster is given the corresponding pseudo-label of that cluster. 
	Thereafter, we could obtain the pseudo-label set $ Y_{sub} = \{ y_{sub_1}, ... , y_{sub_i}, ... \} $,
	where $ y_{sub_i}$ is the pseudo-label of the source context $ c_{sub_i} $. 
	Ultimately, we leverage these pseudo-labels $ Y_{sub} $ of source contexts 
	to fine-tune the PLM with a classifier MLP jointly using the Cross Entropy loss function. 

	It is noted that pseudo-labels generated in the aforementioned way are not perfect. 
	Some source contexts whose corresponding NPs refer to the same entity may not have the same pseudo-label, 
	since the coverage of the collected seed pairs is very limited and many real synonymous NPs do not appear as a seed pair in the collected seed pair set. 
	In order to yield high-quality context embeddings, 
	we require superior pseudo-labels of source contexts for fine-tuning the PLM to further refine its output context embeddings.
	
	To achieve this goal, we propose an iterative clustering procedure to transfer the prior knowledge from limited seed pairs to other pairs, 
	to generate more accurate pseudo-labels, which could be regarded as a training data optimization strategy in terms of data quality. 
	Specifically, in each iteration, we first cluster source contexts via the hierarchical agglomerative clustering (HAC) algorithm 
	based on their context embeddings output by the PLM. 
	Here, we have the assumption that synonymous NPs referring to the same entity may occur in semantically similar source contexts. 
	Consequently, source contexts grouped into the same cluster are expected to mention the same entity and thus should be assigned the same pseudo-label. 
	Then, we yield new pseudo-labels of source contexts based on the new clustering result. 
	Finally, we fine-tune the PLM with a classifier MLP by leveraging these newly generated pseudo-labels. 
	After fine-tuning, PLM is expected to output better context embeddings, 
	which are applied to a new round of iteration to establish a better clustering result and then better pseudo-labels. 
	It can be seen that this iterative clustering procedure performs clustering and fine-tuning alternately to enhance the quality of context embeddings iteratively. 
	Moreover, clustering is the key step to transfer the synonymous relationship of seed pairs to other pairs, 
	via grouping semantically similar source contexts that may mention the same entity into the same cluster, so that they are assigned the same pseudo-label. 
	It is also worth mentioning that we predict the number of clusters for the hierarchical agglomerative clustering algorithm 
	using the Log-Jump algorithm (i.e., Algorithm \ref{alg2}) introduced in Section \ref{sec:inverse-jump} in a data-driven way as well.
	\begin{table}[]
		\caption{Statistics of the used OKB data sets. }
		\vspace{-4.5mm}
		\label{data sets}
		\centering
		\resizebox{0.42\textwidth}{!}{%
			\begin{tabular}{ccccc}
				\hline
				\textit{\textbf{Data set}}    & \textit{\textbf{\#Gold entities}} & \textit{\textbf{\#NPs}} & \textit{\textbf{\#RPs}} & \textit{\textbf{\#OIE triples}} \\ \hline
				\hline
				ReVerb45K   & 7.5k            & 15.5k & 22k       & 45k       \\
				NYTimes2018 & /               & 10.6k & 14k       & 34k       \\
				OPIEC59K    & 18.4k           & 22.8k & 17k       & 59k     \\ \hline
			\end{tabular}%
		}
	\end{table}
	\vspace{-1mm}
	\section{Experimental Study}
	\subsection{Experimental Setting}
	\textbf{Data sets}. 
	Statistics of the three real-world OKB data sets used in the experiments are shown in Table \ref{data sets}.
	We present below a brief summary of each data set.
	
	$\bullet$ ReVerb45K. 
	This OKB data set contains 45k OIE triples extracted by ReVerb \cite{fader2011identifying} from the source text in Clueweb09 \cite{callan2009clueweb09}.
	All NPs in these OIE triples are linked to their corresponding Freebase entities via an entity linking tool \cite{gabrilovich2013facc}, 
	and each entity has at least two aliases occurring as NPs. 
	Unfortunately, this automatic linking process inevitably produces some mistakes. 
	For example, the subject \textit{Google} of the OIE triple \textit{<Google, just bought, Pyra Labs> } and 
	the subject \textit{Yahoo} of the OIE triple \textit{<Yahoo, has relationships with, MSN> } are both linked to the same entity, which is obviously wrong. 
	To mitigate this issue, we manually amended such mistakes existing in 947 OIE triples.
	
	$\bullet$ NYTimes2018. 
	This OKB data set contains 34k OIE triples which are extracted by Standford OIE \cite{angeli2015leveraging} over 1500 articles from nytimes.com in 2018, 
	and all NPs in these OIE triples are not linked to any CKB.
	
	$\bullet$ OPIEC59K. 
	This OKB data set contains 59k OIE triples, 
	distilled from the OPIEC-Linked corpus \cite{gashteovski2019opiec} that contains 5.8M OIE triples. 
	The OPIEC-Linked corpus is extracted by MinIE \cite{gashteovski2017minie} from English Wikipedia articles 
	and contains only those OIE triples in which both NPs (i.e., subject and object) 
	have Wikipedia links. 
	Wikipedia links are anchor texts existing in Wikipedia articles and provided by editors to link NPs in Wikipedia articles with 
	their corresponding Wikipedia entities, which could be regarded as gold disambiguation links. 
	Based on these Wikipedia links, we could group NPs that are linked to the  same Wikipedia entity into the same cluster. 
	Like ReVerb45K, each subject of OIE triples in this data set has at least two aliases. 
	It is worthy to notice that this is an OKB canonicalization data set with the most accurate annotation so far to the best of our knowledge. 
	
	For these three OKB data sets, no training set is given.
	For ReVerb45K, we leverage the triples associated with 20\% selected Freebase entities of ReVerb45K as the validation set, 
	and the rest triples of ReVerb45K as the test set following the previous study \cite{vashishth2018cesi}.  
	For OPIEC59K, we select 10\% OIE triples as the validation set, and the rest triples are regarded as the test set. 
	In this experiment, we use the validation set to search the optimal parameters of semi-supervised baseline methods, 
	and the test set to evaluate the performance of all the methods including both semi-supervised and unsupervised ones. 
	Specially, for the NYTimes2018 data set which is not linked to any CKB, we randomly sample 100 non-singleton NP groups and manually label them 
	as the ground truth for NP canonicalization like the previous study \cite{lin2019canonicalization}.
	For evaluating RP canonicalization on each of the three data sets,
	we randomly sample 35 non-singleton RP groups and manually label them as the ground truth, which is the same as the previous study \cite{lin2019canonicalization}. 
	
	\noindent\textbf{Evaluation measures}. 
	Following the previous OKB canonicalization studies \cite{galarraga2014canonicalizing, vashishth2018cesi, lin2019canonicalization, dash2021open, liu2021joint}, 
	we utilize macro, micro, and pairwise metrics for evaluating the performance of OKB canonicalization methods. 
	Due to the limited space, we omit the detailed computing methods of these metrics and you could refer to 
	\cite{galarraga2014canonicalizing, vashishth2018cesi, lin2019canonicalization, dash2021open, liu2021joint} for more details. 
	Macro, micro, and pairwise metrics evaluate results from different perspectives. 
	To give an overall evaluation of each method, we calculate the average of macro F1, micro F1, and pairwise F1 as \textbf{average F1}, 
	which is a standard comprehensive metric for the task of OKB canonicalization.
	
	\noindent\textbf{Setting details}. 
	For all baselines, we perform grid search over hyperparameter spaces and report their results under the best performing setting. 
	To instantiate our framework CMVC, we need to choose a specific KEM and PLM to work with our framework. 
	In the experiment, we choose the well-known TransE \cite{bordes2013translating} and BERT \cite{kenton2019bert} as the KEM and PLM respectively, which have achieved satisfactory results. 
	We use these two simple models to give prominence to the effects of our framework.
	Note that the performance of choosing different KEMs and PLMs is not the focus of this paper and left for future exploration.
	For the fact view, all input embeddings are initialized via Crawl embeddings, 
	which is a common pre-trained word embeddings by using fastText \cite{bojanowski2017enriching} trained on Common Crawl\footnote{\href{https://commoncrawl.org/2017/06}{https://commoncrawl.org/2017/06}}.
	Specifically, for an NP which contains multiple words, we average the Crawl embeddings of all single words as its embedding for simplicity. 
	The learning rate and the margin hyperparameter are set to $0.0001$ and $12$ respectively.
	For the context view, the learning rate is set to $0.005$. 
	For the multi-view clustering, the number of iterations and the tolerance are set to $10$ and $10^{-4}$ respectively. 
	For the Log-Jump algorithm, we initialize the input embeddings via Crawl embeddings as well. 
	All experiments are implemented by MindSpore Framework\footnote{https://www.mindspore.cn/en}. 
	We make the datasets and the source code used in this paper publicly available for future research\footnote{\href{https://github.com/Yang233666/cmvc}{https://github.com/Yang233666/cmvc}}.
	\begin{table*}[t]
		\caption{Performance on the NP canonicalization task. }
		\vspace{-4.5mm}
		\label{np_all}
		\centering
		\resizebox{0.934\textwidth}{!}{%
			\begin{tabular}{c|ccc|c|ccc|c|ccc|c}
				\hline
				\multirow{2}{*}{\textit{\textbf{Method}}} & \multicolumn{4}{c|}{\textit{\textbf{ReVerb45K}}}                                  & \multicolumn{4}{c|}{\textit{\textbf{NYTimes2018}}}                                  & \multicolumn{4}{c}{\textit{\textbf{OPIEC59K}}}  \\ \cline{2-13}
				& \textit{Macro F1} & \textit{Micro F1} & \textit{Pairwise F1} & \textit{\textbf{Average F1}} & \textit{Macro F1} & \textit{Micro F1} & \textit{Pairwise F1} & \textit{\textbf{Average F1}} & \textit{Macro F1} & \textit{Micro F1} & \textit{Pairwise F1} & \textit{\textbf{Average F1}} \\
				\hline
				\hline
				Morph Norm \cite{vashishth2018cesi}                                & 0.627                       & 0.558                    & 0.334                & 0.506                       & 0.471                    & 0.658                & 0.643                       & 0.590                    & 0.476                & 0.222             & 0.186                & 0.294                            \\ 
				Text Similarity \cite{lin2019canonicalization}                     & 0.625              			& 0.566                    & 0.394                & 0.528                       & 0.581                    & 0.796                & 0.658                       & 0.678                    & 0.480                & 0.228             & 0.192                & 0.300                            \\ 
				IDF Token Overlap \cite{galarraga2014canonicalizing}               & 0.603                       & 0.551                    & 0.338                & 0.497                       & 0.551                    & 0.612                & 0.527                       & 0.563                    & 0.457                & 0.225             & 0.190                & 0.290                            \\ 
				Attribute Overlap \cite{galarraga2014canonicalizing}               & 0.621                       & 0.558                    & 0.342                & 0.507                       & 0.538                    & 0.593                & 0.561                       & 0.564                    & 0.474                & 0.226             & 0.187                & 0.295                            \\ 
				CESI \cite{vashishth2018cesi}                                      & 0.640                       & 0.855                    & 0.842                & 0.779                       & 0.586                    & 0.842                & 0.778                       & 0.735                    & 0.328                & 0.807             & 0.667                & 0.600                            \\ 
				SIST \cite{lin2019canonicalization}                                & /                           & /                        & /                    & /                           & 0.675           & 0.816                & 0.838                       & 0.776                    & /                    & /                 & /                    & /                                \\ 
				CUVA \cite{dash2021open}                                      & \textbf{0.682}                       & 0.872                    & 0.878                & 0.810                       & /                    & /  & /                       & /                    & 0.128                & 0.789             & 0.686                & 0.534                            \\ 
				JOCL$_{cano}$ \cite{liu2021joint}                                      & 0.537                       & 0.854                    & 0.823                & 0.738                       & \textbf{0.876}                    & 0.865  & 0.459                       & 0.733                    & 0.465                & 0.790             & 0.776                & 0.677                            \\ 
				\hline
				CMVC              & 0.662              & \textbf{0.881}           & \textbf{0.893}		   & \textbf{0.812}              & 0.635           		   & \textbf{0.874}       & \textbf{0.893}              & \textbf{0.800}           & \textbf{0.521}       & \textbf{0.909}    & \textbf{0.878}        & \textbf{0.769}                   \\ \hline
			\end{tabular}%
		}
		\vspace{-4mm}
	\end{table*}
	\vspace{-2mm}
	\subsection{Effectiveness Study}
	\vspace{-1mm}
	\subsubsection{NP canonicalization}
	All baselines are listed as follows.
	
	$\bullet$ Morph Norm \cite{fader2011identifying}
	applies several simple normalization operations (e.g., removing tenses, pluralization, and capitalization) over NPs 
	and groups the same NPs after normalization into a group.
	
	$\bullet$ Text Similarity \cite{lin2019canonicalization}
	calculates the Jaro-Winkler similarity \cite{winkler1999state} between two NPs and employs the HAC method to cluster them.
	
	$\bullet$ IDF Token Overlap \cite{galarraga2014canonicalizing}
	measures the similarity between two NPs based on the inverse document frequency (IDF) of their tokens, and leverages the HAC method for clustering.
	
	$\bullet$ Attribute Overlap \cite{galarraga2014canonicalizing}
	leverages the Jaccard similarity of attributes between two NPs for NP canonicalization and clusters NPs via HAC. 
	
	$\bullet$ CESI \cite{vashishth2018cesi}
	is a deep learning based method for OKB canonicalization, which learns the embeddings of NPs and RPs 
	by leveraging relational information in the fact triples and side information, 
	and then clusters the learned embeddings to obtain canonicalized NP (RP) groups via HAC.
	
	$\bullet$ SIST \cite{lin2019canonicalization}
	is an unsupervised method that leverages knowledge embedded in the source text to cluster NPs and RPs jointly for OKB canonicalization. 
	
	$\bullet$ CUVA \cite{dash2021open}
	is a semi-supervised method for OKB canonicalization, which uses variational deep autoencoders to jointly learn both the embeddings and cluster assignments 
	by leveraging relational information in the fact triples and side information. 
	
	$\bullet$ JOCL$_{cano}$ \cite{liu2021joint}
	is an OKB canonicalization framework based on factor graph model, which leverages diverse signals including word embedding, 
	PPDB \cite{pavlick2015ppdb}, AMIE \cite{galarraga2013amie} and transitive relation signals. 
	
	The experimental results of all the methods for NP canonicalization are shown in Table \ref{np_all}.
	Except JOCL$_{cano}$ \cite{liu2021joint}, all the baseline results over the NYTimes2018 data set are directly taken from SIST \cite{lin2019canonicalization}. 
	As we have amended some mistakes in the ReVerb45K data set, ReVerb45K used in this experiment is a little different from it in SIST \cite{lin2019canonicalization}. 
	Therefore, we executed all the baselines over the new version of ReVerb45K as well as the newly constructed OPIEC59K. 
	We fail to obtain the experimental results of some baselines over some data sets, since the source codes obtained from their authors do not work well over these data sets. 
	
	Overall, it can be seen from Table \ref{np_all} that our proposed framework CMVC consistently outperforms all competitive baselines 
	in terms of average F1 over the three data sets.
	The four simple baselines (i.e., Morph Norm, Text Similarity, IDF Token Overlap, and Attribute Overlap) perform poorly over all these three data sets. 
	This may be due to the fact that they mainly rely on the surface forms of NPs and often fail to deal with the cases when NPs with different surface forms refer to the same entity 
	and NPs with similar surface forms refer to different entities. 
	The four recent advanced baselines (i.e., CESI, SIST, CUVA and JOCL$_{cano}$) yield much better performance than the four simple baselines. 
	To be specific, CESI improves the quality of the NP canonicalization via learning NP embeddings using relational information in the fact view 
	and various side information. 
	SIST exceeds CESI on NYTimes2018 by leveraging knowledge embedded in the source context where the OIE fact triple is extracted. 
	CUVA achieves satisfactory performance on ReVerb45K, but performs poorly on OPIEC59K, probably because the performance of variational deep autoencoders is not stable. 
	The performance of JOCL$_{cano}$ is limited as it mainly leverages string and word embedding similarities of NPs. 
	Compared with all these baselines each of which only leverages the knowledge from a single view, 
	our framework CMVC surpasses all of them by integrating the complementary knowledge from both views, 
	which validates the effectiveness of CMVC in NP canonicalization. 
	\vspace{-2.5mm}
	\begin{table}[t]
		\caption{Performance on the RP canonicalization task. }
		\vspace{-4.5mm}
		\label{rp_can}
		\centering
		\resizebox{0.38\textwidth}{!}{%
			\begin{tabular}{cccc|c}
				\hline
				\textit{\textbf{Method}} & \textit{Macro F1} & \textit{Micro F1} & \textit{Pairwise F1} & \textit{\textbf{Average F1}} \\
				\hline
				\hline
				\multicolumn{4}{c|}{\textit{\textbf{ReVerb45K}}}                                                                                                   \\
				AMIE \cite{galarraga2013amie}                     & 0.735                      & 0.863                      & 0.735                         & 0.777                        \\
				PATTY \cite{nakashole2012patty}                   & 0.782                      & 0.872                      & 0.802                         & 0.818                        \\
				CESI \cite{vashishth2018cesi}                     & \textbf{0.923}             & 0.842                      & 0.620                         & 0.795                        \\
				JOCL$_{cano}$ \cite{liu2021joint}                 & 0.918                      & 0.836                      & 0.614                         & 0.789                        \\
				SIST \cite{lin2019canonicalization}               & 0.875                      & 0.872                      & 0.845                         & 0.864                        \\
				CMVC                                              & 0.853                      & \textbf{0.928}             & \textbf{0.856}                & \textbf{0.879}               \\ \hline
				\multicolumn{4}{c|}{\textit{\textbf{NYTimes2018}}}                                                                                                  \\
				PATTY \cite{nakashole2012patty}                   & 0.775                      & 0.802                      & 0.617                         & 0.731                        \\
				JOCL$_{cano}$ \cite{liu2021joint}                 & \textbf{0.885}                      & 0.885                      & 0.522                         & 0.764       \\
				SIST \cite{lin2019canonicalization}               & 0.853             & 0.844                      & 0.722                         & 0.806                        \\
				CMVC                                              & 0.766                      & \textbf{0.905}             & \textbf{0.781}                & \textbf{0.817}               \\ \hline
				\multicolumn{4}{c|}{\textit{\textbf{OPIEC59K}}}                                                                                                    \\
				AMIE \cite{galarraga2013amie}                     & 0.595                      & 0.800                      & 0.631                         & 0.675                        \\
				CESI \cite{vashishth2018cesi}                     & \textbf{0.699}             & 0.752                      & 0.628                         & 0.693                        \\
				JOCL$_{cano}$ \cite{liu2021joint}                 & 0.622                      & 0.775                      & 0.724                         & 0.707                        \\
				CMVC                                              & 0.542                      & \textbf{0.854}             & \textbf{0.770}                & \textbf{0.722}               \\ \hline
				
			\end{tabular}%
		}
	\end{table}
	\subsubsection{RP canonicalization}
	In addition to CESI \cite{vashishth2018cesi}, JOCL$_{cano}$ \cite{liu2021joint} and SIST \cite{lin2019canonicalization}, 
	we add AMIE \cite{galarraga2013amie} and PATTY \cite{nakashole2012patty} as baselines on the task of RP canonicalization. 
	
	$\bullet$ AMIE \cite{galarraga2013amie}
	judges whether two given RPs represent the same semantc meaning by learning Horn rules and requires NPs canonicalized already. 
	Therefore, we only perform AMIE on Reverb45K and OPIEC59K, both of which contain gold NP canonicalization results. 
	If there is a rule $p_i \rightarrow p_j$ output by AMIE, we consider RPs $p_i$ and $p_j$ to be synonymous, and put them into a cluster. 
	
	$\bullet$ PATTY \cite{nakashole2012patty}
	puts OIE triples with the same pairs of NPs as well as RPs belonging to the same synset in PATTY into one cluster.
	
	Experimental results for RP canonicalization are shown in Table \ref{rp_can}.
	It can be seen that CMVC surpasses all baselines in terms of average F1 on all these three data sets.
	AMIE obtains unsatisfactory performance, which may be attributed to the fact that it only covers very few RPs, which leads to most RPs discarded.
	CESI shows superiority over AMIE, since it employs the knowledge embedded in the fact view to cluster RPs with the same semantics.
	PATTY performs well on ReVerb45K since it leverages the synset of each RP for clustering. 
	JOCL$_{cano}$ performs well on OPIEC59K, but performs unsatisfactory over the other two data sets. 
	SIST further improves the performance by utilizing the knowledge embedded in the context view. 
	Compared with SIST, CMVC promotes by over 1 percentage in terms of average F1
	on both ReVerb45K and NYTimes2018 via jointly leveraging the knowledge from the fact view and the context view, 
	indicating the superiority of CMVC in RP canonicalization.
	
	\begin{table}[t]
		\caption{Performance of different variants in CMVC. }
		\vspace{-4.5mm}
		\label{np_ablation}
		\centering
		\resizebox{0.47\textwidth}{!}{%
			\begin{tabular}{ccccc|c}
			\hline
			\multicolumn{2}{c}{\textit{\textbf{Method}}}                                    & \textit{Macro F1} & \textit{Micro F1} & \textit{Pairwise F1} & \textit{\textbf{Average F1}} \\ \hline
			\multicolumn{2}{c}{Seed Pairs}                                                  & 0.308             & 0.900             & 0.872                & 0.693                        \\ \hline
			\multicolumn{1}{c|}{\multirow{2}{*}{Fact View}}    & CMVC$_{fact}-$DAO       & 0.331             & 0.842             & 0.752                & 0.641                        \\
			\multicolumn{1}{c|}{}                              & CMVC$_{fact}$              & 0.516             & 0.888             & 0.815                & 0.739                        \\ \hline
			\multicolumn{1}{c|}{\multirow{2}{*}{Context View}} & CMVC$_{cnt}-$ICP        & 0.020             & 0.817             & 0.806                & 0.547                        \\
			\multicolumn{1}{c|}{}                              & CMVC$_{cnt}$               & 0.340             & 0.905             & \textbf{0.879}       & 0.708                        \\ \hline
			\multicolumn{1}{c|}{\multirow{2}{*}{Multi-View}}   & CMVC$-$CH                & 0.518             & 0.894             & 0.846                & 0.752                        \\
			\multicolumn{1}{c|}{}                              & CMVC                       & \textbf{0.521}    & \textbf{0.909}    & 0.878                & \textbf{0.769}               \\ \hline
			\end{tabular}%
		}
	\end{table}
	
	\vspace{-1.5mm}
	\subsection{Ablation Study}
	\vspace{-1mm}
	To examine the effectiveness of our framework CMVC in combining the complementary knowledge from both views, 
	we remove the multi-view clustering algorithm described in Section \ref{sec:mvc} from CMVC and make two views work alone. 
	We present the performance of three variants, namely, CMVC$_{fact}$ (i.e., CMVC working on the fact view alone), 
	CMVC$_{cnt}$ (i.e., CMVC working on the context view alone), 
	and CMVC (i.e., the whole framework) on OPIEC59K for NP canonicalization in Table \ref{np_ablation}. 
	In addition, to verify the crucial importance of our proposed training data optimization 
	strategy in each view, we remove data augmentation operator (DAO) described in Section \ref{sec:dao} from the variant CMVC$_{fact}$ 
	and remove iterative clustering procedure (ICP) described in Section \ref{sec:icp} from the variant CMV$C_{cnt}$. 
	Furthermore, we remove the CH index described in Section \ref{sec:mvch} from the variant CMVC to verify its importance. 
	We show the experimental results of these three variants in Table \ref{np_ablation} as well as another variant called Seed Pairs 
	that just utilizes the collected seed pairs to generate the canonicalization result. 
	The seed pair collection process is described in Appendix \ref{sec:seed_pairs}. 
	Note that the four variants (i.e., CMVC$_{fact}-$DAO, CMVC$_{fact}$, CMVC$_{cnt}-$ICP and CMVC$_{cnt}$) that output view-specific embeddings 
	are combined with the HAC method to generate the final canonicalization result.
	
	From the experimental results in Table \ref{np_ablation}, we can see that 
	(1) CMVC surpasses CMVC$_{fact}$ and CMVC$_{cnt}$ by 3.0 and 6.1 percentages in terms of average F1, respectively, 
	which validates the point that our proposed multi-view clustering algorithm could indeed mutually reinforce the clustering of view-specific embeddings 
	extracted from each view and harness the complementary knowledge from both views for better canonicalization; 
	(2) compared with CMVC$_{fact}-$DAO (CMVC$_{cnt}-$ICP), CMVC$_{fact}$ (CMVC$_{cnt}$) promotes by an average F1 of 9.8 (16.1) percentages. 
	This confirms that the proposed training data optimization strategy in each view boosts the quality of 
	the learned view-specific embeddings significantly and thus enhances the canonicalization performance obviously; 
	(3) compared with CMVC$-$CH, CMVC promotes by an average F1 of 1.7 percentage, 
	which validates that it is meaningful to consider the different clustering qualities of different views in multi-view clustering rather than treating them equally; 
	(4) in comparison to Seed Pairs, CMVC$_{fact}$ (CMVC$_{cnt}$) improves by an average F1 of 4.6 (1.5) percentages, 
	indicating that both views effectively transfer prior knowledge from seed pairs to other pairs via high-quality encoding. 
	In summary, each component in our proposed framework CMVC has a positive contribution to the canonicalization performance, 
	and when all the components are consolidated together in CMVC, it yields the best performance. 
	\begin{figure*}[t] 
		\subfigcapskip=-10pt
		\vspace{-8mm}
		\begin{minipage}[t]{\textwidth}
			\begin{minipage}[t]{\textwidth}
				\begin{figure}[H]
					\centering
					\subfigure[Fact view.]{ 
						\begin{minipage}[h]{.3\textwidth}
							\centering
							\includegraphics[width=1\textwidth]{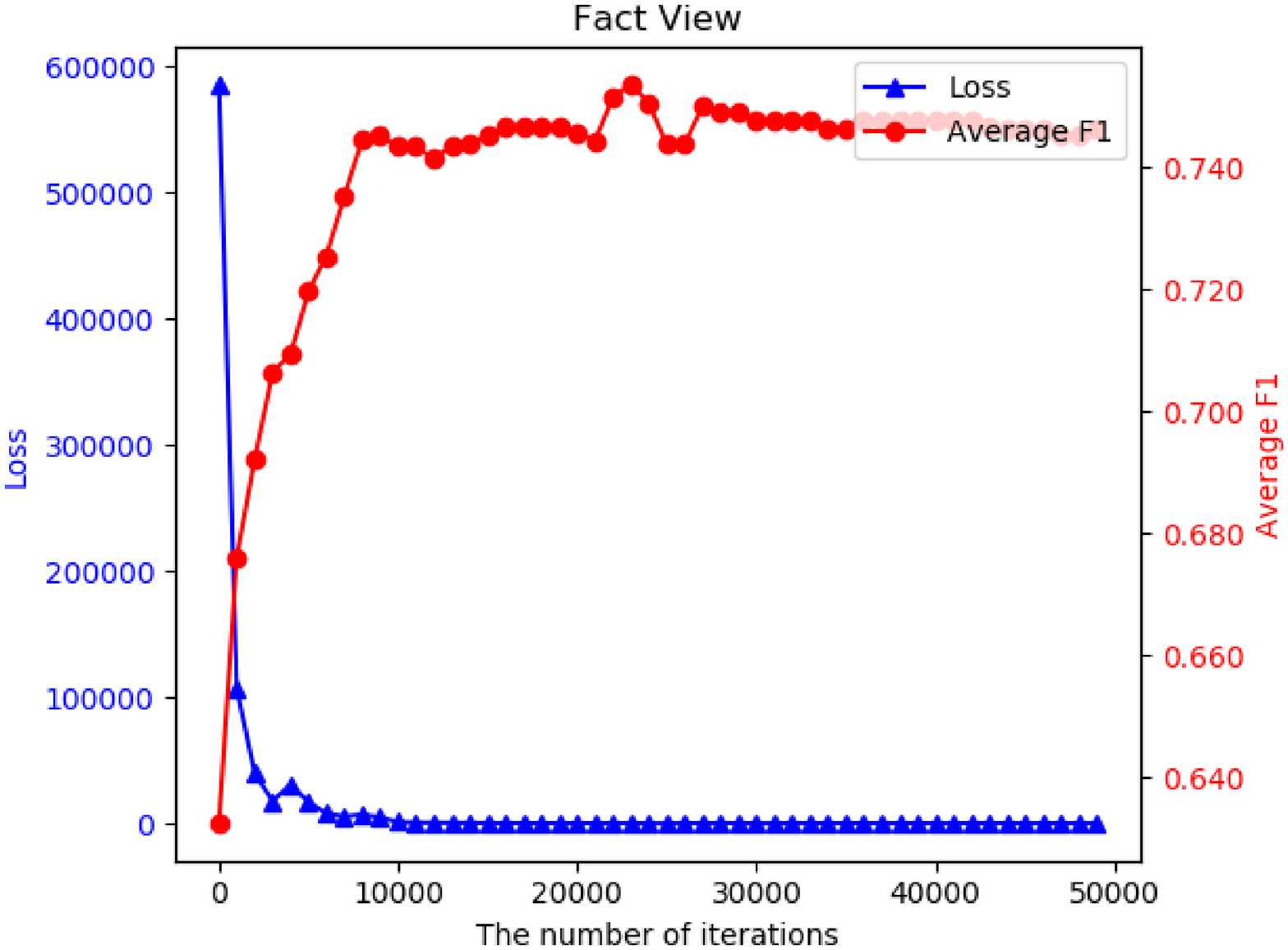}
							\label{fig-2-1}
						\end{minipage}
					}
					\subfigure[Context view.]{    
						\begin{minipage}[h]{.3\textwidth}
							\centering
							\includegraphics[width=1\textwidth]{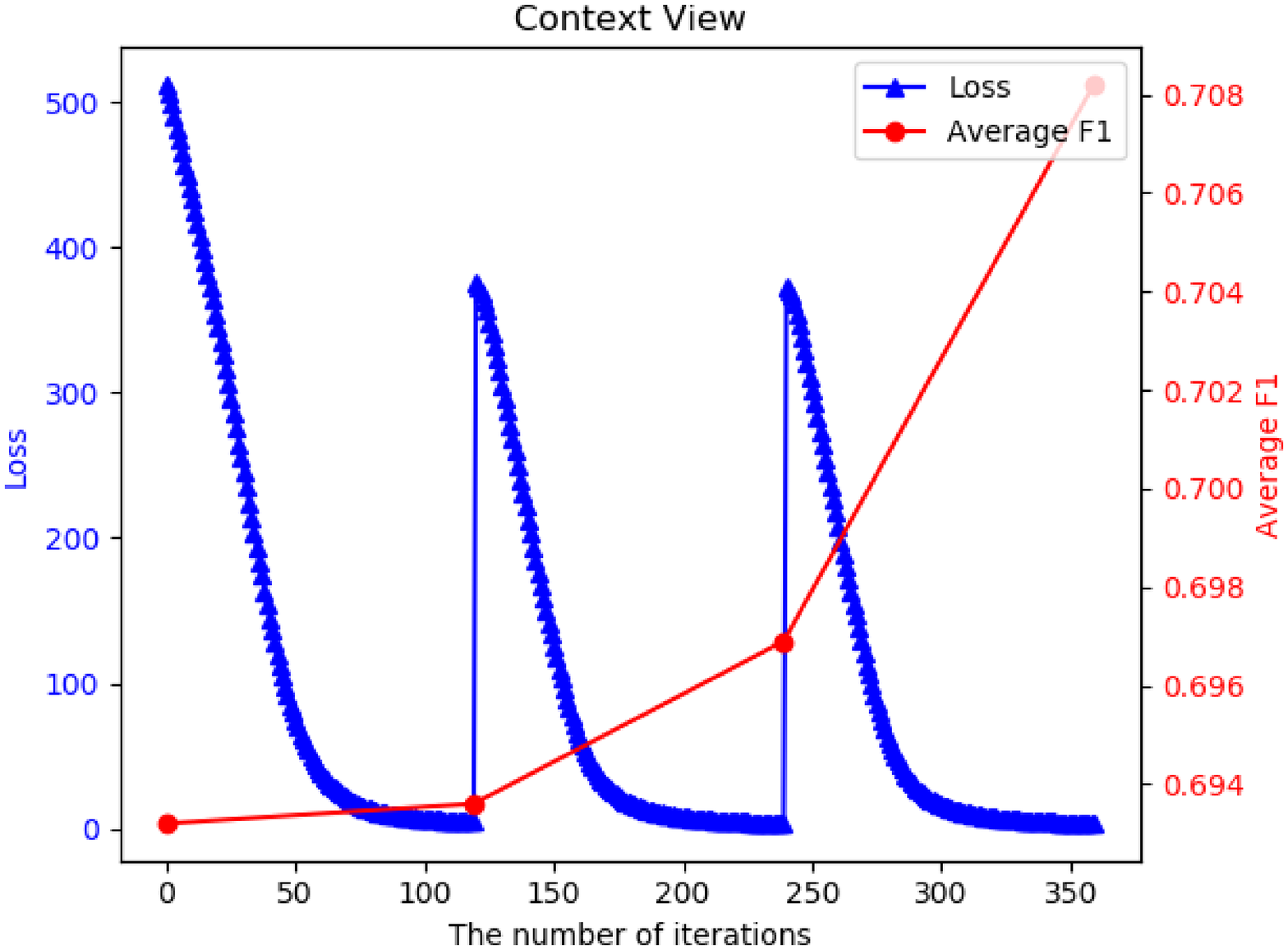}
							\label{fig-2-2}
						\end{minipage}
					}
					\subfigure[CMVC.]{    
						\begin{minipage}[h]{.3\textwidth}
							\centering
							\includegraphics[width=1\textwidth]{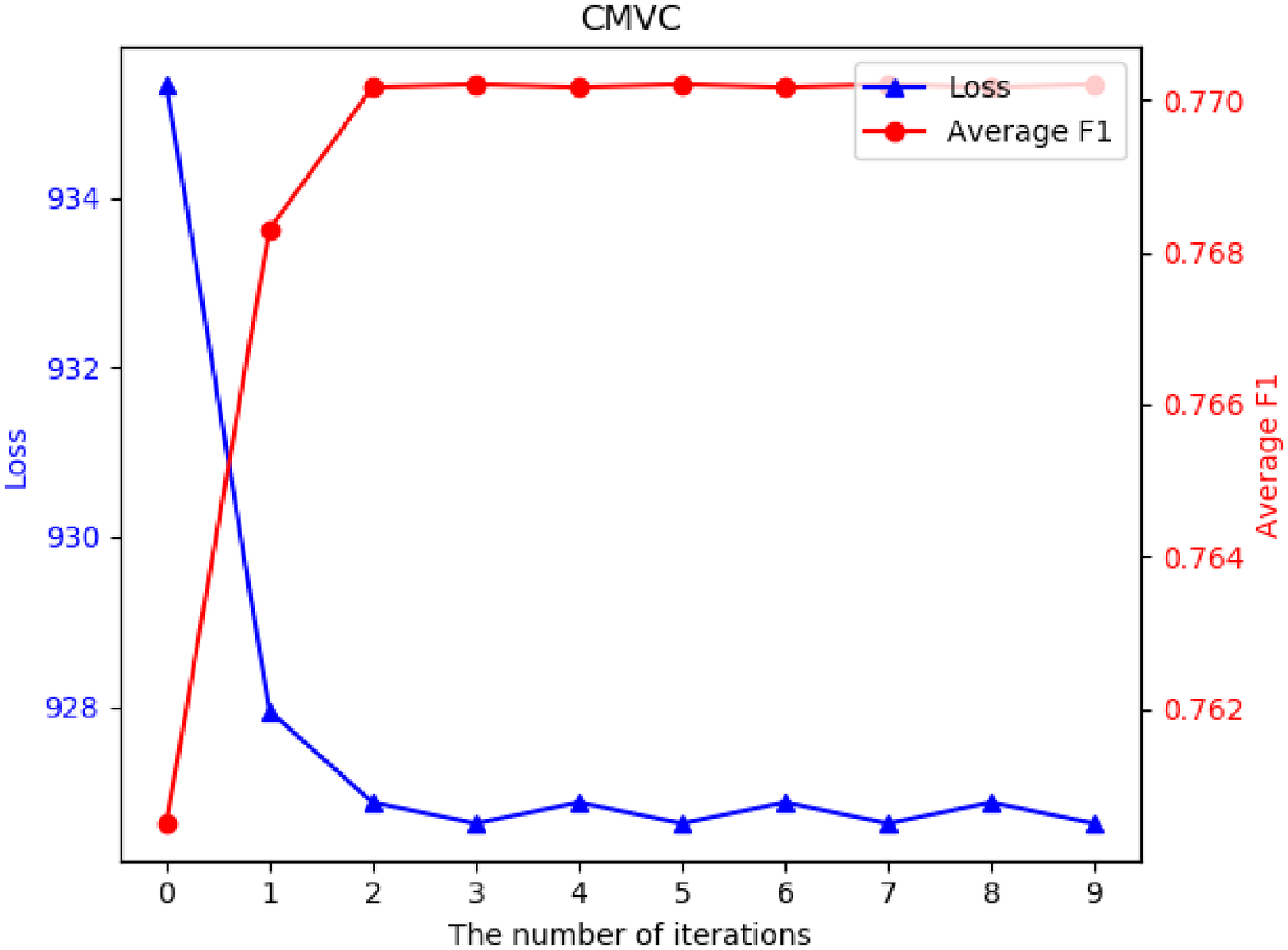}
							\label{fig-2-3}
						\end{minipage}
					}
					\vspace{-6mm}
					\caption{Convergence study. }
					\vspace{-6mm}
				\end{figure}
			\end{minipage}
		\end{minipage}
		\label{figure_2}
	\end{figure*}

	\vspace{-2mm}
	\subsection{Convergence Study}
	To investigate the convergence and effectiveness of our proposed framework CMVC, we show how the loss and average F1 of the fact view, 
	the context view and the whole framework CMVC respectively change with respect to the number of iterations 
	on OPIEC59K in Figure 2. 
	From this figure, we can draw the following observations: 
	(1) the loss decreases rapidly with the increase of the number of iterations in the fact view, the context view, and CMVC, 
	demonstrating that they could achieve a very rapid convergence; 
	(2) the average F1 promotes rapidly as the number of iterations increases in these two single views and CMVC; 
	(3) the increasing speed of the average F1 of the fact view and CMVC slows down as the number of iterations increases. 
	\begin{table}[t]
		\caption{Performance on the cluster number prediction task. }
		\vspace{-4.5mm}
		\label{predict_k_result}
		\centering
		\resizebox{0.43\textwidth}{!}{%
			\begin{tabular}{ccc|ccc}
				\hline
				\textit{\textbf{Method}}     & \textit{\textbf{AR}}     & \textit{\textbf{ARE}}   & \textit{\textbf{Method}}          & \textit{\textbf{AR}}     & \textit{\textbf{ARE}}   \\ \hline
				\hline
				AIC \cite{mehrjou2016improved}        & 6.154  & 0.456 & MPC \cite{rajesh1996fuzzy}             & 6.769  & 0.522 \\
				BIC \cite{zhao2008knee}        & 15.077 & 0.724 & PC \cite{james1973cluster}              & 7.385  & 0.533 \\
				CH Index \cite{calinski1974dendrite}   & 6.462  & 0.422 & PI \cite{bensaid1996validity}              & 5.385  & 0.436 \\
				CE \cite{bezdek1975classe}         & 15.385 & 0.736 & PBMF \cite{pakhira2004validity}            & 5.692  & 0.436 \\
				CWB \cite{rezaee1998new}        & 14.154 & 0.655 & PCAES \cite{wu2005cluster}           & 4.692  & 0.391 \\
				DB Index \cite{davies1979cluster}   & 5.462  & 0.440 & RLWY Index \cite{ren2016self}      & 15.385 & 0.736 \\
				Dunn Index \cite{dunn1973fuzzy} & 10.154 & 0.644 & Rezaee \cite{rezaee2010cluster}          & 16.000 & 0.753 \\
				Knee-point \cite{salvador2004knee} & 2.538  & 0.341 & SIL Index \cite{rousseeuw1987silhouettes}       & 6.462  & 0.550 \\
				FS Index \cite{fukuyama1989ANM}   & 3.846  & 0.389 & Slope Statistic \cite{fujita201427} & 5.923  & 0.553 \\
				FHV \cite{rajesh1996fuzzy}        & 13.923 & 0.710 & XB Index \cite{xie1991validity}        & 15.923 & 0.750 \\
				HV Index \cite{halkidi2001clustering}   & 16.000 & 0.753 & Xu Index \cite{xu1997bayesian}        & 11.923 & 0.611 \\
				I Index \cite{maulik2002performance}    & 5.538  & 0.414 & ZXF Index \cite{zhao2009sum}       & 5.077  & 0.422 \\
				LL \cite{gupta2018fast}         & 7.923  & 0.558 & Jump \cite{sugar2003findingtn}            & 8.538  & 0.415 \\
				LML \cite{gupta2018fast}        & 7.364  & 0.560 & Log-Jump        & \textbf{1.769}  & \textbf{0.217} \\ \hline
			\end{tabular}%
		}
	\end{table}
	\vspace{-2mm}
	\subsection{Effect Analysis of Data-Driven Parameter Prediction}
	To validate the effectiveness of our proposed Log-Jump algorithm (introduced in Section \ref{sec:inverse-jump}) in predicting the parameter (i.e., the number of clusters), 
	we compare it with twenty seven existing cluster number prediction methods on a total of thirteen real-world data sets. 
	We obtain six data sets from \cite{gupta2018fast} and five data sets from OpenML \cite{OpenML2013}, and show the number of input embeddings, 
	the dimension of the input embedding and the number of clusters for each data set in Table \ref{predict_k_data set} of Appendix \ref{sec:experiments_details}. 
	In addition to these eleven data sets, we leverage two OKB canonicalization data sets (i.e., ReVerb45K and OPIEC59K) 
	both of which have been annotated for NP canonicalization. 
	Note that the numbers of clusters in these two OKB data sets are much larger than the other eleven data sets, which brings a new challenge. 
	To give an overall evaluation of each cluster number prediction method, besides the average of rank (AR) over all data sets used in \cite{gupta2018fast}, 
	we calculate the average of relative error (ARE) as well. 
	The metric of relative error is defined as $\frac{\left\vert k - k^{*} \right\vert}{k^{*}} \label{re}$, 
	where $k$ and $k^*$ denote the predicted number of clusters and the gold number of clusters of a data set, respectively.
	
	From the results shown in Table \ref{predict_k_result}, it can be seen that our proposed method Log-Jump 
	significantly outperforms all the twenty seven baselines in terms of AR and ARE, exhibiting the superiority 
	of Log-Jump for predicting the number of clusters. 
	Moreover, our proposed Log-Jump algorithm could be applied to not only the OKB canonicalization task addressed in this paper, 
	but also other clustering tasks that need to estimate the number of clusters. 
	The detailed experimental results of all methods over each data set are shown in Tables \ref{predict_k_all_rank_result} and \ref{predict_k_all_relative_error_result} of Appendix \ref{sec:experiments_details}. 
	
	\vspace{-3mm}
	\section{RELATED WORK}
	Two aspects of research are related to our work: OKB canonicalization and multi-view clustering, which are introduced as follows. 
	
	For the task of OKB canonicalization, previous methods can be divided into two categories: 
	semi-supervised methods \cite{galarraga2014canonicalizing, wu2018towards, vashishth2018cesi, dash2021open, liu2021joint} and unsupervised method \cite{lin2019canonicalization}.
	Our proposed framework CMVC belongs to the latter without the requirement of any manually annotated label. 
	In addition, according to the view of knowledge leveraged for OKB canonicalization, previous methods could also be classified into two types: 
	fact view based methods \cite{galarraga2014canonicalizing, wu2018towards, vashishth2018cesi, dash2021open, liu2021joint} and context view based method \cite{lin2019canonicalization}. 
	Specifically, the first OKB canonicalization model \cite{galarraga2014canonicalizing} performs the HAC method over manually-defined features, 
	such as IDF token overlap and attribute overlap, to cluster synonymous NPs, and then clusters RPs by leveraging AMIE \cite{galarraga2013amie}. 
	Based on this canonicalization model \cite{galarraga2014canonicalizing}, 
	FAC \cite{wu2018towards} proposes a more efficient graph-based clustering method, 
	which utilizes pruning and bounding techniques to reduce similarity computation. 
	CESI \cite{vashishth2018cesi} clusters the embeddings of NPs (RPs), 
	which are learned by leveraging a KB embedding model based on the fact view and various side information 
	(i.e., Entity Linking, PPDB, WordNet, IDF token overlap, morph normalization, AMIE, and KBP), to obtain canonicalized NP (RP) groups. 
	SIST \cite{lin2019canonicalization} leverages knowledge from the original source text (i.e., the context view), 
	to cluster NPs and RPs jointly using an efficient clustering method. 
	CUVA \cite{dash2021open} uses variational deep autoencoders to jointly learn both embeddings and cluster assignments in a semi-supervised way. 
	JOCL$_{cano}$ \cite{liu2021joint} is based on factor graph model by leveraging diverse signals including word embedding, 
	PPDB, AMIE and transitive relation signals. 
	All aforementioned works only leverage the knowledge from a single view. 
	Beyond that, our framework CMVC harnesses the complementary knowledge from both views effectively for OKB canonicalization. 
	
	Multi-view clustering aims to provide more accurate and stable partitions than single view clustering 
	by considering complementary information embedded in multiple views,
	which has achieved prominent success in diverse tasks, such as graph clustering \cite{ fan2020one2multi} and dialog intent induction \cite{ perkins2019dialog}.  
	To the best of our knowledge, this paper is the first work that leverages multi-view clustering to investigate the task of OKB canonicalization. 
	
	\vspace{-1.5mm}
	\section{CONCLUSION}
	\vspace{-0.5mm}
	The complementarity between the fact view and the context view is vital to the task of OKB canonicalization, but previous works ignore it. 
	In this paper, we propose a novel unsupervised framework CMVC that integrates the complementary knowledge delivered by both views 
	via a multi-view CH K-Means algorithm by considering their different clustering qualities. 
	In order to further enhance the canonicalization performance, we propose a training data optimization strategy in terms of 
	data quantity and data quality respectively in each particular view to refine the learned view-specific embeddings in an iterative manner. 
	To demonstrate the effectiveness of CMVC, we conduct extensive experiments over three real-world OKB data sets, 
	and the experimental results show that our framework surpasses all baselines in terms of average F1. 
	\vspace{-2mm}
	\begin{acks}
	This work was supported in part by National Natural Science Foundation of China (No. U1936206), YESS by CAST (No. 2019QNRC001), and CAAI-Huawei MindSpore Open Fund. 
	\end{acks}
	\vspace{-2mm}
	
	\bibliographystyle{ACM-Reference-Format}
	\bibliography{sample-base}
	
	\clearpage
	\normalsize
	\appendix
	
	\section{Pseudo-code of Multi-View CH K-Means Algorithm\label{sec:multi-view_ch}}
	\vspace{-3.5mm}
	\begin{algorithm}[htbp]
		\renewcommand{\algorithmicrequire}{\textbf{Input:}}
		\renewcommand{\algorithmicensure}{\textbf{Output:}}
		\caption{Multi-View CH K-Means Algorithm}
		\label{alg1}
		\begin{algorithmic}[1]
			\REQUIRE A set of NPs $S = \{ sub_1, ... , sub_i, ...  \} $ and their view-specific embeddings of two views $ \{(\bm{sub}_{1}^{1}, \bm{sub}_{1}^{2}), ..., (\bm{sub}_{i}^{1}, \bm{sub}_{i}^{2}), ...\}$, the number of iterations $T$, the tolerance $tol$, the number of clusters $K$. 
			\STATE Initialization: randomly initialize the cluster center embeddings $ \pmb{\xi}^{(2)} $ of view 2. 
			\STATE $\bm{E}$-$\bm{step}$ in view $2$: compute the clustering result $ \pi^{(2)} $ by Formula~(\ref{mvc_e})
			\STATE $ t = 0 $
			\REPEAT	
			\FOR{$v=1$ to $2$}{
				\STATE $t \leftarrow t + 1$
				\STATE $\bm{M}$-$\bm{step}$ in view $v$: compute the cluster center embeddings $ \pmb{\xi}^{(v)} $ by Formula~(\ref{mvc_m}) based on the clustering result $ \pi^{(\bar{v})}$ from the other view $\bar{v}$ 
				\STATE $\bm{E}$-$\bm{step}$ in view $v$: compute the clustering result $ \pi^{(v)} $ by Formula~(\ref{mvc_e})
			}
			\ENDFOR
			\STATE Compute the loss function $\mathcal{L}_{mvc}$ by Formula~(\ref{mvc_loss})
			\UNTIL $ t == T $ or $\mathcal{L}_{mvc}$ < $tol$
			\STATE Compute consensus means $\bm{m}^{(1)}_j$ and $\bm{m}^{(2)}_j$ by Formula \eqref{mvc_mean}
			\STATE Compute CH indexes ${CH}^{(1)}$ and ${CH}^{(2)}$ by Formula \eqref{ch}
			\STATE Compute the final clustering result $\hat{\pi}$ by Formula \eqref{mvc_pi}
			\ENSURE the final clustering result $ \hat{\pi} $
		\end{algorithmic}  
	\end{algorithm}
	\vspace{-4.5mm}
	\section{Supplementary Materials for the Log-Jump Algorithm}
	\subsection{Proof of the Log-Jump Algorithm\label{A1}}
	To give a theoretical justification for the Log-Jump algorithm based on information theoretic ideas, 
	inspired by the proof in \cite{sugar2003findingtn}, we give Theorem 1 and its proof as follows:
	
	\textbf{Theorem 1} \emph{Suppose that the distribution of the input X is a mixture of G Gaussian clusters with equal priors and common covariance $\Gamma_{p}$. 
		Let $\Delta \sqrt{p}$ be the minimum cosine distance between cluster means after standardizing the space by multiplying by $\Gamma_{p}^{-1/2}$. 
		Thereafter, for $ K < G $}
	\vspace{-0.7mm}
	\begin{equation}
	\lim_{p \to \infty} d_{K} = c \quad (\frac{p \Delta^2}{9G}W < c \le 1) \label{Theorem 1}
	\end{equation}
	\vspace{-0.7mm}
	where $W = 1 - \frac{6^4 V_X}{{( \Delta^2 - 36 )}^2}$, 
	$ V_X = \min\limits_{j \in \{ 1,..,K \} } \frac{1}{p} \sum_{l = 1} ^{p} { ( X_{jl} - \mu_{jl} )^2 }$, 
	$\mu_{j}$ is the cluster center embedding of $j$-th cluster, and $\Delta > 6$. 
	
	\textbf{Proof}. According to Theorem 2 in \cite{sugar2003findingtn}, 
	\vspace{-0.7mm}
	\begin{equation}
	d_{K} \ge \frac{p \Delta^2}{9G}W, 
	\end{equation}
	\vspace{-0.7mm}
	As for $G > p$ and ${p \to \infty}$, thus $\frac{p \Delta^2}{9G}W > 0$. $d_{K}$ is calculated by cosine distance, thus $ d_{K} \le 1 $. 
	Let $d_{K} = c$ when ${p \to \infty}$, then we could obtain Formula \eqref{Theorem 1}. 
	
	Theorem 1 implies that for large enough $p$ and $G$, $K<G$, then $d_{K}^{-p/2} \approx c, (\frac{p \Delta^2}{9G}W \le c \le 1)$. 
	While for $K>G$, thus $d_{K}^{-p/2} \varpropto k^{p} \approx K $. 
	It is easy to obtain $d_{K}$ as follows:
	\vspace{-0.7mm}
	\begin{equation}
	d_{K} = \begin{cases}
	{( \frac{G}{aK} )} ^{2/p}, & \text {$K > G$} \quad (0 < a < 1) \\
	c, &\text{$K \le G$} \quad (\frac{p \Delta^2}{9G}W \le c \le 1) 
	\end{cases}  \label{d_k}
	\end{equation}
	\vspace{-0.7mm}
	Next, we could calculate the Log-Jump measure $LJ_K$ based on the logarithm of the distortion as follows: 
	\vspace{-0.7mm}
	\begin{equation}
	LJ_{K} = \log d_{K+1} - \log d_{K} = \begin{cases}
	c - c = 0, & \text {K < G} \\
	\log \frac{1}{a} > 0, &\text{K = G}  \\
	\frac{2}{p} \log \frac{K}{K+1} < 0, &\text{K > G}
	\end{cases}  \label{log_d_k}
	\end{equation}
	\vspace{-0.7mm}
	Finally, according to Formula \eqref{log_d_k}, the number of clusters $G$ can be estimated as follows: 
	\vspace{-0.7mm}
	\begin{equation}
	G = \arg\max\limits_{K} { LJ_{K} }
	\end{equation}
	
	\vspace{-1.5mm}
	\subsection{Heuristic Method to Generate a Small Candidate Range of Possible $K$\label{A2}}
	In this section, we introduce a heuristic method to generate a small candidate range of possible $K$ for the sake of efficiency. 
	Otherwise, we have to test each possible $K$ from $1$ to $n$ in Algorithm \ref{alg2}, which is time-consuming. 
	
	For traditional clustering tasks, the number of clusters is usually small, and the candidate range of possible $K$ could be set as 
	$ \left[ 1, \left \lfloor \frac{\sqrt{n}}{2} \right \rfloor \right] $. 
	For example, if the number of input embeddings $n = 100$, the candidate range of possible $K$ changes from 
	the original range $ \left[1, 100\right] $ to a small range $ \left[1, 5\right] $. 
	
	Considering the large number of clusters for the task of OKB canonicalization, 
	we need to extend $\log d_{K+1} - \log d_{K}$ to $\log d_{K+gap} - \log d_{K}$, where $gap$ is a positive integer. 
	Specifically, we change the step size from $1$ to $gap$, which could be calculated as follows:
	\begin{equation}
	gap = 10^{len(string(n)) - 2}, \label{inverse-jump-max} 
	\end{equation}
	where function $string(\cdot)$ maps an integer $n$ to its string style, and $len(\cdot)$ calculates the length of a string.
	For example, if the number of input embeddings $n = 20000$, then the $gap$ will be $10^{len(``20000")-2}=10^{5-2} = 1000$. 
	If the number of input embeddings $n > 10000$, the number of clusters may be large, and thus we generate the candidate range of possible $K$ as 
	$ \left[ 4gap, 9gap \right] $. 
	Otherwise, the number of clusters may be small, and then we generate the candidate range of possible $K$ as 
	$ \left[ 2gap, 9gap \right] $. 
	For example, if the number of input embeddings $n = 20000$, then the step size changes from $1$ to $1000$, 
	and the candidate range of possible $K$ changes from the original range $ \left[1, 20000\right] $ to a small range $ \left[4000, 9000\right] $. 
	In order to get a more accurate result, we could set $gap = gap / 10$, and perform the next round of iteration to predict the number of clusters 
	according to the rules introduced above. 
	
	\vspace{-1.5mm}
	\section{Seed pair collection\label{sec:seed_pairs}}
	We collect seed pairs automatically from three external resources 
	(i.e., mention entity dictionary \cite{Shen2012linden, shen2021entity}, search engine, and AMIE \cite{galarraga2013amie}) without the need of any human involvement. 
	
	$\bullet$ Mention entity dictionary \label{sec:ep} contains information about possible mapping entities of various mentions 
	and their corresponding prior mapping probabilities calculated based on the count information in Wikipedia. 
	Given an NP, we look up the dictionary and obtain its most likely mapping entity which has the maximum prior mapping probability. 
	For two NPs, if their most likely mapping entities are the same, they will form a seed pair. 
	
	$\bullet$ Search engine would return a collection of relevant Web pages for a given query. We query an NP (a RP) 
	via Bing\footnote{\href{https://www.bing.com/}{https://www.bing.com/}} and collect the URLs of the returned Web pages in the first ten search result pages. 
	For two NPs (RPs), if the Jaccard similarity of their corresponding URL collections is greater than a threshold, they will form a seed pair. 
	In the experiment, this threshold is set to 0.015.
	
	$\bullet$ AMIE\label{sec:amie} judges whether two given RPs are synonymous by learning Horn rules, and it could be used to generate seed pairs of RPs directly. 
	It requires the input data to be semi-canonicalized (i.e., NPs are canonicalized already), so we normalize NPs morphologically and then apply AMIE over the NP-canonicalized OKB. 
	
	\section{Detailed experimental results on cluster number prediction \label{sec:experiments_details}} 
	For the task of cluster number prediction, we show the statistics of the eleven data sets in Table \ref{predict_k_data set} and the detailed experimental results 
	of all methods over all data sets in Tables \ref{predict_k_all_rank_result} and \ref{predict_k_all_relative_error_result}. 
	\vspace{-6.5mm}
	\begin{table*}[]
		\caption{The rank metrics of all methods over all data sets on the cluster number prediction task. }
		\vspace{-4.5mm}
		\label{predict_k_all_rank_result}
		\centering
		\resizebox{0.872\textwidth}{!}{%
			\begin{tabular}{ccccccccccccccc}
				\hline
				\textit{\textbf{Method}}       & \textit{\textbf{Echocardiogram}} & \textit{\textbf{Iris}} & \textit{\textbf{Seeds}} & \textit{\textbf{Wine}} & \textit{\textbf{Colon}} & \textit{\textbf{Prostate}} & \textit{\textbf{Mice}} & \textit{\textbf{Abalone}} & \textit{\textbf{Cnae-9}} & \textit{\textbf{Vowel}} & \textit{\textbf{Synthetic}} & \textit{\textbf{OPIEC59K}} & \textit{\textbf{ReVerb45K}} & {\textit{\textbf{AR}}}                                                  \\ \hline
				\hline
				AIC \cite{mehrjou2016improved}   & 1                                & 1                      & 1                       & 1                      & 1                              & 1                                 & 11                             & 14                        & 11                       & 9                       & 10                                  & 10                         & 9                           & 6.154                                               \\
				BIC \cite{zhao2008knee} & 19                               & 20                     & 18                      & 16                     & 1                              & 1                                 & 19                             & 21                        & 18                       & 21                      & 19                                  & 12                         & 11                          & 15.077                                              \\
				CH Index \cite{calinski1974dendrite}             & 1                                & 1                      & 1                       & 16                     & 1                              & 1                                 & 11                             & 3                         & 11                       & 9                       & 2                                   & 13                         & 14                          & 6.462                                               \\
				CE \cite{bezdek1975classe}         & 19                               & 20                     & 18                      & 16                     & 1                              & 1                                 & 19                             & 21                        & 18                       & 21                      & 19                                  & 13                         & 14                          & 15.385                                              \\
				CWB \cite{rezaee1998new}         & 19                               & 20                     & 18                      & 16                     & 1                              & 1                                 & 19                             & 21                        & 18                       & 21                      & 19                                  & 4                          & 7                           & 14.154                                              \\
				DB Index \cite{davies1979cluster}           & 1                                & 1                      & 1                       & 1                      & 1                              & 1                                 & 11                             & 14                        & 11                       & 9                       & 10                                  & 6                          & 4                           & 5.462                                               \\
				Dunn Index \cite{dunn1973fuzzy}                     & 1                                & 1                      & 26                      & 16                     & 1                              & 27                                & 3                              & 5                         & 4                        & 4                       & 10                                  & 21                         & 13                          & 10.154                                              \\
				Knee-point \cite{salvador2004knee}              & 1                                & 1                      & 1                       & 1                      & 1                              & 1                                 & 7                              & 5                         & 3                        & 7                       & 2                                   & 2                          & 1                           & 2.538                                               \\
				FS Index \cite{fukuyama1989ANM}          & 1                                & 1                      & 1                       & 1                      & 1                              & 1                                 & 7                              & 8                         & 7                        & 7                       & 2                                   & 5                          & 8                           & 3.846                                               \\
				FHV \cite{rajesh1996fuzzy}              & 19                               & 1                      & 18                      & 16                     & 1                              & 1                                 & 19                             & 21                        & 18                       & 21                      & 19                                  & 13                         & 14                          & 13.923                                              \\
				HV Index \cite{halkidi2001clustering}     & 19                               & 20                     & 18                      & 16                     & 1                              & 1                                 & 19                             & 21                        & 18                       & 21                      & 19                                  & 21                         & 14                          & 16.000                                              \\
				I Index \cite{maulik2002performance}                        & 1                                & 1                      & 1                       & 1                      & 1                              & 1                                 & 11                             & 9                         & 8                        & 9                       & 2                                   & 13                         & 14                          & 5.538                                               \\
				LL \cite{gupta2018fast}                      & 19                               & 1                      & 1                       & 1                      & 1                              & 1                                 & 19                             & 14                        & 18                       & 9                       & 2                                   & 7                          & 10                          & 7.923                                               \\
				LML \cite{gupta2018fast}                & 19                               & 1                      & 1                       & 1                      & 1                              & 1                                 & 19                             & 14                        & 18                       & 4                       & 2                                   & /                          & /                           & 7.364                                               \\
				MPC \cite{rajesh1996fuzzy} & 1                                & 1                      & 1                       & 1                      & 1                              & 1                                 & 11                             & 9                         & 8                        & 9                       & 10                                  & 21                         & 14                          & 6.769                                               \\
				PC \cite{james1973cluster}          & 1                                & 1                      & 1                       & 1                      & 1                              & 1                                 & 11                             & 14                        & 11                       & 9                       & 10                                  & 21                         & 14                          & 7.385                                               \\
				PI \cite{bensaid1996validity}                & 1                                & 1                      & 1                       & 1                      & 1                              & 1                                 & 11                             & 14                        & 11                       & 9                       & 10                                  & 7                          & 2                           & 5.385                                               \\
				PBMF \cite{pakhira2004validity}                           & 1                                & 1                      & 1                       & 16                     & 1                              & 1                                 & 6                              & 9                         & 5                        & 4                       & 2                                   & 13                         & 14                          & 5.692                                               \\
				PCAES \cite{wu2005cluster}                          & 1                                & 1                      & 1                       & 1                      & 1                              & 1                                 & 1                              & 9                         & 1                        & 9                       & 2                                   & 19                         & 14                          & 4.692                                               \\
				RLWY Index \cite{ren2016self}          & 19                               & 20                     & 18                      & 16                     & 1                              & 1                                 & 19                             & 21                        & 18                       & 21                      & 19                                  & 13                         & 14                          & 15.385                                              \\
				Rezaee \cite{rezaee2010cluster}                         & 19                               & 20                     & 18                      & 16                     & 1                              & 1                                 & 19                             & 21                        & 18                       & 21                      & 19                                  & 21                         & 14                          & 16.000                                              \\
				SIL Index \cite{rousseeuw1987silhouettes}               & 1                                & 1                      & 1                       & 1                      & 1                              & 1                                 & 7                              & 9                         & 8                        & 9                       & 10                                  & 21                         & 14                          & 6.462                                               \\
				Slope Statistic \cite{fujita201427}                & 1                                & 1                      & 1                       & 1                      & 1                              & 1                                 & 7                              & 5                         & 18                       & 9                       & 10                                  & 11                         & 11                          & 5.923                                               \\
				XB Index \cite{xie1991validity}                 & 19                               & 20                     & 18                      & 16                     & 1                              & 1                                 & 19                             & 21                        & 18                       & 21                      & 19                                  & 20                         & 14                          & 15.923                                              \\
				Xu Index \cite{xu1997bayesian}                       & 1                                & 20                     & 26                      & 16                     & 1                              & 27                                & 3                              & 1                         & 5                        & 1                       & 19                                  & 21                         & 14                          & 11.923                                              \\
				ZXF Index \cite{zhao2009sum}                      & 1                                & 1                      & 1                       & 1                      & 1                              & 1                                 & 11                             & 14                        & 11                       & 9                       & 10                                  & 3                          & 2                           & 5.077                                               \\
				Jump \cite{sugar2003findingtn}                    & 1                                & 20                     & 26                      & 16                     & 1                              & 1                                 & 3                              & 1                         & 11                       & 1                       & 19                                  & 7                          & 4                           & 8.538                                               \\ \hline
				Log-Jump            & 1                                & 1                      & 1                       & 1                      & 1                              & 1                                 & 1                              & 4                         & 1                        & 3                       & 1                                   & 1                          & 6                           & {\textbf{1.769}}                                    \\ \hline
			\end{tabular}%
		}\vspace{-3mm}
	\end{table*}
	\vspace{-6.5mm}
	
	\vspace{-6.5mm}
	\begin{table*}[]
		\caption{The relative error metrics of all methods over all data sets on the cluster number prediction task. }
		\vspace{-4.5mm}
		\label{predict_k_all_relative_error_result}
		\centering
		\resizebox{0.872\textwidth}{!}{%
			\begin{tabular}{ccccccccccccccc}
				\hline
				\textit{\textbf{Method}}       & \textit{\textbf{Echocardiogram}} & \textit{\textbf{Iris}} & \textit{\textbf{Seeds}} & \textit{\textbf{Wine}} & \textit{\textbf{Colon}} & \textit{\textbf{Prostate}} & \textit{\textbf{Mice}} & \textit{\textbf{Abalone}} & \textit{\textbf{Cnae-9}} & \textit{\textbf{Vowel}} & \textit{\textbf{Synthetic}} & \textit{\textbf{OPIEC59K}} & \textit{\textbf{ReVerb45K}} & \textit{\textbf{ARE}} \\ \hline
				\hline
				AIC \cite{mehrjou2016improved}   & 0.333                            & 0.333                  & 0.333                   & 0.333                  & 0.000                   & 0.000                      & 0.750                  & 0.929                     & 0.778                    & 0.818                   & 0.667                       & 0.490                      & 0.170                       & 0.456                           \\
				BIC \cite{zhao2008knee} & 0.667                            & 0.667                  & 0.667                   & 0.667                  & 0.500                   & 0.500                      & 0.875                  & 0.964                     & 0.889                    & 0.909                   & 0.833                       & 0.776                      & 0.506                       & 0.724                           \\
				CH Index \cite{calinski1974dendrite}             & 0.333                            & 0.000                  & 0.000                   & 0.667                  & 0.000                   & 0.000                      & 0.750                  & 0.214                     & 0.778                    & 0.818                   & 0.500                       & 0.796                      & 0.632                       & 0.422                           \\
				CE \cite{bezdek1975classe}         & 0.667                            & 0.667                  & 0.667                   & 0.667                  & 0.500                   & 0.500                      & 0.875                  & 0.964                     & 0.889                    & 0.909                   & 0.833                       & 0.796                      & 0.632                       & 0.736                           \\
				CWB \cite{rezaee1998new}         & 0.667                            & 0.667                  & 0.667                   & 0.667                  & 0.500                   & 0.500                      & 0.875                  & 0.964                     & 0.889                    & 0.909                   & 0.833                       & 0.265                      & 0.110                       & 0.655                           \\
				DB Index \cite{davies1979cluster}           & 0.333                            & 0.333                  & 0.333                   & 0.333                  & 0.000                   & 0.000                      & 0.750                  & 0.929                     & 0.778                    & 0.818                   & 0.667                       & 0.347                      & 0.094                       & 0.440                           \\
				Dunn Index \cite{dunn1973fuzzy}                     & 0.333                            & 0.333                  & 1.000                   & 0.667                  & 0.500                   & 1.000                      & 0.250                  & 0.821                     & 0.556                    & 0.636                   & 0.667                       & 1.020                      & 0.582                       & 0.644                           \\
				Knee-point \cite{salvador2004knee}              & 0.333                            & 0.333                  & 0.333                   & 0.333                  & 0.000                   & 0.000                      & 0.625                  & 0.821                     & 0.333                    & 0.727                   & 0.500                       & 0.082                      & 0.005                       & 0.341                           \\
				FS Index \cite{fukuyama1989ANM}          & 0.333                            & 0.333                  & 0.333                   & 0.333                  & 0.000                   & 0.000                      & 0.625                  & 0.857                     & 0.556                    & 0.727                   & 0.500                       & 0.306                      & 0.154                       & 0.389                           \\
				FHV \cite{rajesh1996fuzzy}              & 0.667                            & 0.333                  & 0.667                   & 0.667                  & 0.500                   & 0.500                      & 0.875                  & 0.964                     & 0.889                    & 0.909                   & 0.833                       & 0.796                      & 0.632                       & 0.710                           \\
				HV Index \cite{halkidi2001clustering}     & 0.667                            & 0.667                  & 0.667                   & 0.667                  & 0.500                   & 0.500                      & 0.875                  & 0.964                     & 0.889                    & 0.909                   & 0.833                       & 1.020                      & 0.632                       & 0.753                           \\
				I Index \cite{maulik2002performance}                        & 0.000                            & 0.000                  & 0.000                   & 0.333                  & 0.000                   & 0.000                      & 0.750                  & 0.893                     & 0.667                    & 0.818                   & 0.500                       & 0.796                      & 0.632                       & 0.414                           \\
				LL \cite{gupta2018fast}                      & 0.667                            & 0.333                  & 0.333                   & 0.333                  & 0.500                   & 0.500                      & 0.875                  & 0.929                     & 0.889                    & 0.818                   & 0.500                       & 0.367                      & 0.209                       & 0.558                           \\
				LML \cite{gupta2018fast}                & 0.667                            & 0.000                  & 0.333                   & 0.333                  & 0.500                   & 0.500                      & 0.875                  & 0.929                     & 0.889                    & 0.636                   & 0.500                       & /                          & /                           & 0.560                           \\
				MPC \cite{rajesh1996fuzzy} & 0.333                            & 0.333                  & 0.333                   & 0.333                  & 0.000                   & 0.000                      & 0.750                  & 0.893                     & 0.667                    & 0.818                   & 0.667                       & 1.020                      & 0.632                       & 0.522                           \\
				PC \cite{james1973cluster}          & 0.333                            & 0.333                  & 0.333                   & 0.333                  & 0.000                   & 0.000                      & 0.750                  & 0.929                     & 0.778                    & 0.818                   & 0.667                       & 1.020                      & 0.632                       & 0.533                           \\
				PI \cite{bensaid1996validity}                & 0.333                            & 0.333                  & 0.333                   & 0.333                  & 0.000                   & 0.000                      & 0.750                  & 0.929                     & 0.778                    & 0.818                   & 0.667                       & 0.367                      & 0.028                       & 0.436                           \\
				PBMF \cite{pakhira2004validity}                           & 0.000                            & 0.000                  & 0.000                   & 0.667                  & 0.000                   & 0.500                      & 0.375                  & 0.893                     & 0.667                    & 0.636                   & 0.500                       & 0.796                      & 0.632                       & 0.436                           \\
				PCAES \cite{wu2005cluster}                          & 0.333                            & 0.333                  & 0.000                   & 0.000                  & 0.500                   & 0.000                      & 0.000                  & 0.893                     & 0.111                    & 0.818                   & 0.500                       & 0.959                      & 0.632                       & 0.391                           \\
				RLWY Index \cite{ren2016self}          & 0.667                            & 0.667                  & 0.667                   & 0.667                  & 0.500                   & 0.500                      & 0.875                  & 0.964                     & 0.889                    & 0.909                   & 0.833                       & 0.796                      & 0.632                       & 0.736                           \\
				Rezaee \cite{rezaee2010cluster}                         & 0.667                            & 0.667                  & 0.667                   & 0.667                  & 0.500                   & 0.500                      & 0.875                  & 0.964                     & 0.889                    & 0.909                   & 0.833                       & 1.020                      & 0.632                       & 0.753                           \\
				SIL Index \cite{rousseeuw1987silhouettes}               & 0.333                            & 0.333                  & 0.333                   & 0.333                  & 0.500                   & 0.000                      & 0.625                  & 0.893                     & 0.667                    & 0.818                   & 0.667                       & 1.020                      & 0.632                       & 0.550                           \\
				Slope Statistic \cite{fujita201427}                & 0.333                            & 0.333                  & 0.333                   & 0.333                  & 0.500                   & 0.500                      & 0.625                  & 0.821                     & 0.889                    & 0.818                   & 0.667                       & 0.531                      & 0.506                       & 0.553                           \\
				XB Index \cite{xie1991validity}                 & 0.667                            & 0.667                  & 0.667                   & 0.667                  & 0.500                   & 0.500                      & 0.875                  & 0.964                     & 0.889                    & 0.909                   & 0.833                       & 0.980                      & 0.632                       & 0.750                           \\
				Xu Index \cite{xu1997bayesian}                       & 0.333                            & 0.667                  & 1.000                   & 0.667                  & 0.500                   & 1.000                      & 0.250                  & 0.107                     & 0.667                    & 0.273                   & 0.833                       & 1.020                      & 0.632                       & 0.611                           \\
				ZXF Index \cite{zhao2009sum}                      & 0.333                            & 0.333                  & 0.333                   & 0.333                  & 0.000                   & 0.000                      & 0.750                  & 0.929                     & 0.778                    & 0.818                   & 0.667                       & 0.184                      & 0.028                       & 0.422                           \\
				Jump \cite{sugar2003findingtn}                    & 0.333                            & 0.667                  & 1.000                   & 0.667                  & 0.000                   & 0.000                      & 0.250                  & 0.107                     & 0.778                    & 0.273                   & 0.833                       & 0.388                      & 0.094                       & 0.415                           \\ \hline
				Log-Jump            & 0.000                            & 0.333                  & 0.333                   & 0.333                  & 0.000                   & 0.500                      & 0.125                  & 0.464                     & 0.000                    & 0.455                   & 0.167                       & 0.000                      & 0.104                       & \textbf{0.217}                  \\ \hline
			\end{tabular}%
		}\vspace{-3mm}
	\end{table*}
	\vspace{-6.5mm}
	\begin{table}[]
		\caption{Statistics of the data sets for cluster number prediction. }
		\vspace{-4.5mm}
		\label{predict_k_data set}
		\centering
		\resizebox{0.45\textwidth}{!}{%
			\begin{tabular}{cccc|cccc}
				\hline
				\textit{\textbf{Data set}} & \textit{\textbf{\#Emb}} & \textit{\textbf{\#Dim}} & \textit{\textbf{\#Clu}} & \textit{\textbf{Data set}} & \textit{\textbf{\#Emb}} & \textit{\textbf{\#Dim}} & \textit{\textbf{\#Clu}} \\ \hline
				\hline
				Echocardiogram    & 106       & 9     & 3     & Mice Protein      & 552       & 77      & 8       \\
				Iris              & 150       & 4     & 3     & Abalone           & 4177      & 8       & 28        \\
				Seeds             & 210       & 7     & 3     & Cnae-9            & 1080      & 856     & 9         \\
				Wine              & 178       & 13    & 3     & Vowel             & 990       & 12      & 11        \\
				Colon Cancer      & 62        & 2000  & 2     & Synthetic Control & 600       & 60      & 6         \\
				Prostate Cancer   & 102       & 6033  & 2     \\ \hline
			\end{tabular}%
		}
	\end{table}
	
\end{document}